\RequirePackage{fix-cm}
\documentclass[twocolumn]{svjour3}          % twocolumn
\smartqed  % flush right qed marks, e.g. at end of proof
\usepackage{slashbox}
\usepackage{subfigure}
\usepackage{graphicx}
\usepackage{amsmath,amssymb}
\usepackage{multirow}
\usepackage{array}

\newcommand{\tabincell}[2]{\begin{tabular}{@{}#1@{}}#2\end{tabular}}
\begin{document}

\title{Bag of Visual Words and Fusion Methods for Action Recognition: Comprehensive Study and Good Practice%\thanks{Grants or other notes
}
% \subtitle{Do you have a subtitle?\\ If so, write it here}


\author{Xiaojiang Peng  \and
        Limin Wang \and
        Xingxing Wang \and
        Yu Qiao
}


\institute{Xiaojiang Peng \at
              School of Information Sciences and Technology, Southwest Jiaotong University, Chengdu, China \\
              \email{xiaojiangp@gmail.com}
           \and
           Limin Wang \at
              Department of Information Engineering, The Chinese University of Hong Kong, Hong Kong SAR, China. \\
%              Tel.: (+852)\ 2609-8260\\
%              Fax: (+852)\ 2603-5032\\
              \email{07wanglimin@gmail.com}
           \and
           Xingxing Wang \at
              School of Electrical and Electronic Engineering, Nanyang Technological University, Singapore. \\
              \email{wangxingxing.hz@gmail.com}
           \and
           Yu Qiao \at
               Shenzhen Institutes of Advanced Technology, Chinese Academy of Sciences, Shenzhen, China. \\
               \email{yu.qiao@siat.ac.cn}
}


\maketitle

\begin{abstract}
Video based action recognition is one of the important and challenging problems in computer vision research. Bag of Visual Words model (BoVW) with local features has become the most popular method and obtained the state-of-the-art performance on several realistic datasets, such as the HMDB51, UCF50, and UCF101. BoVW is a general pipeline to construct a global representation from a set of local features, which is mainly composed of five steps: (i) feature extraction, (ii) feature pre-processing, (iii) codebook generation, (iv) feature encoding, and (v) pooling and normalization. Many efforts have been made in each step independently in different scenarios and their effects on action recognition is still unknown. Meanwhile, video data exhibits different views of visual pattern, such as static appearance and motion dynamics. Multiple descriptors are usually extracted to represent these different views. Fusing these multiple descriptors is crucial for boosting the final performance of action recognition system. Many feature fusion methods have been developed in other areas and their influence on action recognition has never been investigated before. This paper aims to provide a comprehensive study of all steps in BoVW and different fusion methods, and uncover some good practice to produce a state-of-the-art action recognition system. Specifically, we explore two kinds of local features, ten kinds of encoding methods, eight kinds of pooling and normalization strategies, and three kinds of fusion methods. We conclude that every step is crucial for contributing to the final recognition rate and improper choice in one of the steps may counteract the performance improvement of other steps. Furthermore, based on our comprehensive study, we propose a simple yet effective representation, called \emph{hybrid representation}, by exploring the complementarity of different BoVW frameworks and local descriptors. Using this representation, we obtain the state-of-the-art on the three challenging datasets: HMDB51 ($61.1\%$), UCF50 ($92.3\%$), and UCF101 ($87.9\%$).
\end{abstract}

\section{Introduction}
\label{intro}
Human action recognition \cite{AggarwalR11,TuragaCSU08} has become an important area in computer vision research, whose aim is to automatically classify the  action ongoing in a video. It is one of the challenging problems in computer vision for serval reasons. Firstly, there are large intra-class variations in the same action class, caused by various motion speeds, viewpoint changes, and background clutter. Secondly, the identification of an action class is related to many other high-level visual clues, such as human pose, interacting objects, and scene class. These related problems are very difficult themselves. Furthermore, the determination of temporal extent for an actions is more subjective than a static object, which means there is no precise definition about when an action starts and finishes. Finally, the high dimension and low quality of video data usually add difficulty to developing robust and efficient recognition algorithm.

Early approaches interpret an action as a set of space-time trajectories of 2-dimensional or 3-dimensional points of human joints \cite{Webb81,NiyogiA94,CampbellB95,YacoobB99}. These methods usually need dedicate techniques to detect body parts or track them at each frame. However, the detection and tracking of body part is still an unsolved problem in realistic videos. Recently, recognition methods using local spatiotemporal features \cite{Laptev05,LaptevMSR08,WangKSL13,WangQT14} have become the main stream and obtained the state-of-the-art performance on many datasets \cite{WangS13a}. These methods do not require algorithms to detect human body, which treat the action volume as a rigid 3D-object and extract appropriate features to describe the patterns of each 3D volume. They are robust to background clutter, illumination changes, and noise.

Bag of Visual Words (BoVW) framework with local features and its variants \cite{Wang13,Wu13,Karaman13,Murthy13,PengWCQP13} have dominated the research work of action recognition and showed their effectiveness in the recent THUMOS'13 Action Recognition Challenge \cite{THUMOS13}. As shown in Figure \ref{fig:flowchart}, the pipeline of BoVW for video based action recognition consists of five steps: (i) feature extraction, (ii) feature pre-processing, (iii) codebook generation, (iv) feature encoding, and (v) pooling and normalization. In each step, many efforts have been made and several progress has been obtained. Regarding local features, many successful feature extractors (e.g. STIPs \cite{Laptev05}, Dense Trajectories \cite{WangKSL13}) and descriptors (e.g. HOG \cite{LaptevMSR08}, HOF \cite{LaptevMSR08}, MBH \cite{WangKSL13}) have been designed for representing the visual patterns of cuboid. Feature pre-processing technique mainly de-correlates these descriptors to make the following representation learning more stable. For codebook generation, it aims to describe the local feature space and provide a partition (e.g. $k$-means \cite{Bishop06}) or generative process (e.g. GMMs \cite{Bishop06}) for local descriptor.  Feature encoding is a hot topic in image classification and many alternatives have been developed for effective representation and efficient implementation (see good surveys \cite{ChatfieldLVZ11,HuangWWT14}). Max pooling \cite{YangYGH09} and sum pooling \cite{ZhangMLS07} are usually used to aggregate information from a spatiotemporal region. For normalization methods, typical choices include $\ell_1$-normalization \cite{ZhangMLS07}, $\ell_2$-normalization \cite{WangYYLHG10}, power normalization \cite{PerronninSM10}, and intra normalization \cite{ArandjelovicZ13}. \emph{How to make decision in each step to obtain the best pipeline of BoVW for action recognition} still remains unknown and needs to be extensively explored.

Meanwhile, unlike static image, video data exhibits different views of visual pattern, such as appearance, motion, and motion boundary, and all of them play important roles in action recognition. Therefore, multiple descriptors are usually extracted from a cuboid and each descriptor corresponds to the specific aspect of the visual data \cite{WangKSL13,LaptevMSR08}. BoVW is mainly designed for a single descriptor and ignores the problem of fusing multiple descriptors. Many research works have been devoted to fusing multiple descriptor for boosting performance  \cite{GehlerN09,VedaldiGVZ09,TangYLK13,WangS13a,CaiWPQ14}. Typical fusion methods include descriptor level fusion \cite{LaptevMSR08,WangWQ12}, representation level fusion \cite{Wang13,WangKSL13}, and score level fusion \cite{TangYLK13,MyersNHPNSHKSSS14}. For descriptor level fusion, multiple descriptors from the same cuboid are concatenated as a whole one and fed into BoVW framework. For representation level fusion, the fusion is conducted in the video level, where each descriptor is firstly fed into BoVW framework independently and the resulting global representations are then concatenated to train a final classifier. For score level fusion, each descriptor is separately input into BoVW framework and used to train a recognition classifier. Then the scores from multiple classifiers are fused using arithmetic mean or geometric mean. In general, these fusion methods are developed in different scenarios and adapted for action recognition by different works. \emph{How these fusion methods influence the final recognition of BoVW framework and whether there exists a best one for action recognition} is an interesting question and well worth of a detailed investigation.

Several related study works have been performed about encoding methods for image classification \cite{ChatfieldLVZ11,HuangWWT14} and action recognition \cite{WangWQ12}. But these study works are with image classification task or lacking full exploration of all steps in BoVW framework. Meanwhile, the study work of action recognition \cite{WangWQ12} is limited regarding the evaluation dataset and ignores the influence of fusion methods. This article aims to \emph{provide a comprehensive study of all steps in BoVW and different fusion methods, and uncover some good practice to produce a state-of-the-art action recognition system}. Our work is mainly composed of three parts:

\textbf{Exploration of BoVW.} We place an emphasis on extensively explorations about all components in BoVW pipeline and discovery of useful practice tips. Specifically, we investigate two widely-used local features, namely Space Time Interest Points (STIPs) with HOG, HOF \cite{Laptev05}, and Improved Dense Trajectories (iDTs) with HOG, HOF, MBH \cite{WangS13a}. For feature encoding methods, the current approaches can be roughly classified into three categories: (i) voting based encoding methods, (ii) reconstruction based encoding methods, (iii) super vector based encoding methods. For each type of encoding methods, we choose several representative approaches and totally analyze ten encoding methods. Meanwhile, we explore the relations among these different encoding methods and provide an unified and generative perspective over these encoding methods. We fully explored eight pooling and normalization strategies for each encoding method. From our extensive study of different components in BoVW, server good practice can be concluded:
\begin{itemize}
  \item Dense features with more descriptors are more informative in capturing the content of video data and suitable for action recognition. Meanwhile, dense features may exhibit different properties with sparse features with respect to variations of BoVW such as codebook size and encoding methods.
  \item Data pre-processing is an important step in BoVW pipeline and able to greatly improve the final recognition performance.
  \item Basically, high dimensional representation of super vector is more effective and efficient than the other two types of encoding methods.
  \item Pooling and normalization is a crucial step in BoVW, whose importance may not be highlighted in previous studies. Sum pooling with power $\ell_2$-normalization is the best choice during all the possible combinations.
  \item In above, every step is crucial for contributing to the final recognition rate. Improper choice in one of the steps may counteract the performance improvement of other steps.
\end{itemize}

\textbf{Investigation of Fusion Methods.} As combination of multiple descriptors is very crucial for performance improvement, we also investigate the influence of different fusion methods inour designed action recognition system. Specifically, we study three kinds of fusion methods, namely descriptor level fusion, representation level fusion, and descriptor level fusion. We find that the way different descriptors correlate with each other determines the effectiveness of fusion methods. The performance gain obtained from fusing multiple descriptors mainly owns to their complementarity. We observe that this complementarity is not only with multiple descriptors, but also with multiple BoVW models. Based on this view, we propose a new representation, called \emph{hybrid representation}, combining the outputs of multiple BoVW models of different descriptors. This representation utilizes the benefit of each BoVW and fully considers the complementarity among them. In spite of its simplicity, this representation turns out to be effective for improving final recognition rate.

\textbf{Comparison with the State of the Art.} Guided by the practice tips concluded from our insightful analysis of BoVW variants and feature fusion methods, we design an effective action recognition system using our proposed hybrid representation, and demonstrates its performance on three challenging datasets: HMDB51 \cite{KuehneJGPS11}, UCF50 \cite{ReddyS13}, and UCF101 \cite{SOOMRO12}. Specifically, we leverage the richness and effectiveness of low-level features, design a hybrid super vector, a combination of Fisher vector \cite{PerronninSM10} and SVC-$k$, and resort to representation level fusion to boost final recognition performance. From comparison with other methods, we conclude that our recognition system reaches the state-of-the-art performance on the three datasets, and our hybrid representation acts as a new baseline for further research of action recognition.

The rest of this paper is organized as follows. In Section \ref{sec:method}, we give an detailed description of each step in BoVW framework of action recognition system. Meanwhile, we uncover several useful techniques commonly adopted in these encoding methods, and provide a unified generative perspective over these encoding methods. Then, several fusion methods and a new representation are introduced in Section \ref{sec:fusion}. Finally, we empirically evaluate the BoVW frameworks and fusion methods on three challenging datasets. We analyze these experiment results and uncover good practice for constructing a state-of-the-art action recognition system. We conclude the paper in Section \ref{sec:conclusion}.

\section{Framework of Bag of Visual Words}
\label{sec:method}
\begin{figure*}
  \centerline{\includegraphics[width=\textwidth]{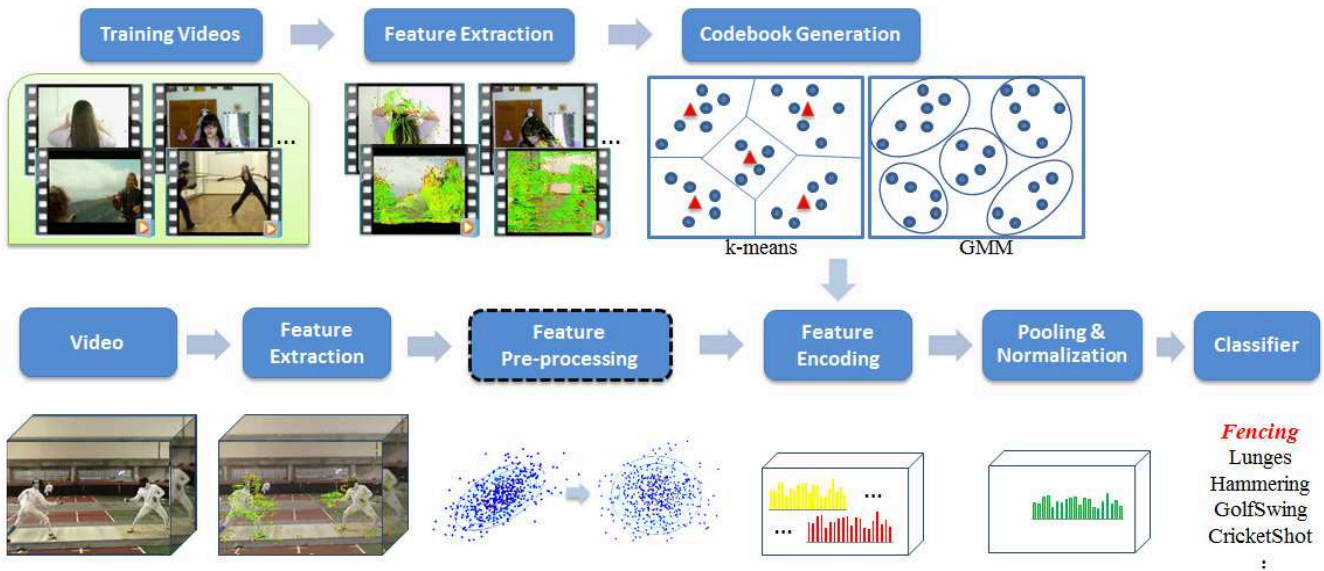}}
  \caption{The pipeline of obtaining Bag of Visual Words (BoVWs) representation for action recognition. It is mainly composed of five steps: (i) feature extraction, (ii) feature pre-processing, (iii) codebook generation, (iv) feature encoding, and (v) pooling and normalization.}
  \label{fig:flowchart}
\end{figure*}

As shown in Figure \ref{fig:flowchart}, the pipeline of Bag of Visual Words (BoVWs) framework consists of five steps: (i) feature extraction, (ii) feature pre-processing, (iii) codebook generation, (iv) feature encoding, and (v) pooling and normalization. Then the global representation is fed into a classifier such as linear SVM for action recognition. In this section, we will give detailed descriptions of the popular technical choices in each step, which are very important for constructing a state-of-the-art recognition system. Furthermore, we summarize several use techniques in these encoding methods and provide a unified generative perspective over these different encoding methods.

\begin{table*}[!htb]
\caption{List of encoding methods and their formulations. The detailed descriptions of these encoding methods can be found in the text.}
\centering
\begin{tabular}{|c|l|l|c|}\hline
Type & Method & Formulation & Dim.\\ \hline
\multirow{5}{*}[-1cm]{\rotatebox{90}{Voting based}} &1. Vector Quantization (VQ) / Hard Voting (HV) & \tabincell{l}{$\mathbf{s}(i) = 1, \ \textrm{if} \ \ i=\textrm{arg}\min_{j}||\mathbf{x} - \mathbf{d}_j||_2^2, $\\$ \textrm{s.t.} \ ||\mathbf{s}||_{0} = 1$ } & $K$\\
\cline{2-4}
& 2. Soft-assignment (SA) / Kernel Codebook Coding (KCB) & $\mathbf{s}(i) = \frac{\exp(-\beta||\mathbf{x}-\mathbf{d}_i||_2^{2})}{\sum_{i=1}^K \exp(-\beta||\mathbf{x} - \mathbf{d}_i||_2^{2})} $ & $K$\\
\cline{2-4}
& 3. Localized Soft Assignment (SA-k) & \tabincell{l}{$ \mathbf{s}(i) = \frac{\exp(-\beta||\mathbf{x}-\mathbf{d}_i||_2^{2})}{\sum_{i=1}^k \exp(-\beta||\mathbf{x} - \mathbf{d}_i||_2^{2})},\ \textrm{if} \ \mathbf{d}_i\in N_{k}(\mathbf{x})$,\\ $ \textrm{s.t.} \ ||\mathbf{s}||_{0} = k $} & $K$ \\
\cline{2-4}
& 4. Salient Coding (SC) & \tabincell{l}{$ \mathbf{s}(i) = \sum_{j=2}^k(\frac{||\mathbf{x}-\mathbf{d}_j||_2-||\mathbf{x}-\mathbf{d}_1||_2}{||\mathbf{x}-\mathbf{d}_j||_2}),\mathbf{d}_j\in N_{k}(\mathbf{x})$,\\$ \textrm{if} \quad i=\textrm{arg}\min_{j}||\mathbf{x} - \mathbf{d}_j||_2^2, \quad \textrm{s.t.} \ ||\mathbf{s}||_{0} = 1$} & $K$\\
\cline{2-4}
& 5. Group Salient Coding (GSC) & \tabincell{l}{$ \mathbf{s}(i) = \max\{\mathbf{v}^k(i)\}, \ \textrm{group size } k=1,...,M$,\\$ \mathbf{v}^k(i)=\sum_{j=1}^{M+1-k}||\mathbf{x}-\mathbf{d}_{k+j}||_2-||\mathbf{x}-\mathbf{d}_k||_2$,\\$ \textrm{if} \ \mathbf{d}_i\in N_{k}(\mathbf{x})$, \ \textrm{s.t.} \ $\|\mathbf{s}\|_0 = 1$ }  & $K$ \\
\hline\hline
\multirow{4}{*}[-0.2cm]{\rotatebox{90}{Reconstruction based}}& 6. Orthogonal Matching Pursuit (OMP) & $\min_{\mathbf{s}}||\mathbf{x}-\mathbf{D} \mathbf{s}||_2^2, \ \textrm{s.t.} \ ||\mathbf{s}||_{0}\le k $ & $K$ \\
\cline{2-4}
& 7. Sparse Coding (SPC) & $ \mathbf{s} = \textrm{arg}\min_{\mathbf{s}}||\mathbf{x-Ds}||_2^2+\lambda ||\mathbf{s}||_1$ & $K$\\
\cline{2-4}
& 8. Locality-constrained Linear Coding (LLC) & \tabincell{l}{$\mathbf{s} = \textrm{arg}\min_{\mathbf{s}}||\mathbf{x-Ds}||_2^2+\lambda||\mathbf{e}\odot\mathbf{s}||_2^2,$\\
$ \textrm{s.t.} \quad \mathbf{1^{\top}s} = 1$} & $K$ \\
\cline{2-4}
& 9. Local Coordinate Coding (LCC) & $\mathbf{s} = \textrm{arg}\min_{\mathbf{s}}||\mathbf{x-Ds}||_2^2+\lambda ||\mathbf{\hat e}\odot|\mathbf{s}|||_1$ & $K$ \\
\hline\hline
\multirow{4}{*}[-1cm]{\rotatebox{90}{Super vector based}}& 10. Local Tangent-based Coding (LTC) & \tabincell{l}{$\mathcal{S}=[\mathbf{s}(i),\mathbf{s}(i)(\mathbf{x}-\mathbf{d}_i)^T\mathbf{U}_i]_{i=1}^K,$ \\
$ \mathbf{U}_i \in \mathbb{R}^{D \times C} \ \textrm{is a projection matrix} $} & $K(1+C)$ \\
\cline{2-4}
& 11. Super Vector Coding (SVC) &  \tabincell{l}{$ \mathcal{S}=[0,\mathbf{0},\cdots,\frac{\alpha\mathbf{s}(i)}{N\sqrt{p_i}},\frac{\mathbf{s}(i)}{N\sqrt{p_i}}(\mathbf{x}-\mathbf{d}_i),\cdots,0,\mathbf{0}]$, \\ $\textrm{where} \ i=\textrm{arg}\min_{j}||\mathbf{x} - \mathbf{d}_j||_2^2$ }& $K(1+D)$\\
\cline{2-4}
& 12. Fisher Vector (FV) & \tabincell{l}{$\mathcal{S} = [\mathcal{G}_{\mu,1}^\mathbf{x},...,\mathcal{G}_{\mu,K}^\mathbf{x},\mathcal{G}_{\sigma,1}^\mathbf{x},...,\mathcal{G}_{\sigma,K}^\mathbf{x}], \textrm{where}$\\
$\mathcal{G}_{\mu,i}^\mathbf{x}=\frac{1}{\sqrt{\pi_i}}\gamma_i(\frac{\mathbf{x}-\mathbf{\mu}_i}{\sigma_i})$,\\
$\mathcal{G}_{\sigma,i}^\mathbf{x}=\frac{1}{\sqrt{2\pi_i}}\gamma_i[\frac{(\mathbf{x}-\mathbf{\mu}_i)^2}{\sigma_i^2}-1]$,\\
$\gamma(i)=\frac{\pi_i \mathcal{N}(\mathbf{x};\mu_i,\Sigma_i)}{\sum_{j=1}^K \pi_j \mathcal{N}(\mathbf{x};\mu_j,\Sigma_j)}$} & $2KD$\\
\cline{2-4}
& 13. Vector of Locally Aggregated Descriptors (VLAD) &  \tabincell{l}{$ \mathcal{S}=[\mathbf{0},\cdots,(\mathbf{x}-\mathbf{d}_i),\cdots,\mathbf{0}]$, \\ $\textrm{where} \ i=\textrm{arg}\min_{j}||\mathbf{x} - \mathbf{d}_j||_2^2$ } & $KD$\\
\hline
\end{tabular}
\label{tab:list}
\end{table*}

\subsection{Feature Extraction}
Low-level local features have become popular in action recognition due to their robustness to background clutter and independence on detection and tracking techniques. These local features are typically divided into two parts: detecting a local region (detector) and describing the detected region (descriptor) \cite{WangUKLS09}. Many feature detectors have been developed such as 3D-Harris \cite{Laptev05}, 3D-Hessian \cite{WillemsTG08}, Cuboid \cite{DollarRCB05}, Dense Trajectories \cite{WangKSL13}, and Improved Dense Trajectories \cite{WangS13a}. These detectors try to select locations and scales in video by maximizing certain kind of function or using dense sampling strategy. To describe the extracted region, several hand-crafted features have been designed such as Histogram of Oriented Gradients (HOG) \cite{LaptevMSR08,WangKSL13}, Histogram of Oriented Flow (HOF) \cite{LaptevMSR08,WangKSL13}, and Motion Boundary Histogram (MBH) \cite{WangKSL13,WangS13a}. Multiple descriptors are usually adopted to represent the local region, each of which corresponds to a certain aspect of visual pattern such as static appearance, motion, and motion boundary.

Among these local features, Space Time Interest Points (STIPs) \cite{Laptev05} and Improved Dense Trajectories (iDTs) \cite{WangKSL13} are widely used due to their easy usages and good performance. STIPs resort to 3D-Harris to extract regions of high motion salience, which resulting a set of sparse interest points. For each interest point, STIPs extracted two kinds of descriptors, namely HOG and HOF. iDTs features are an improved version from Dense Trajectories (DTs), where a set of dense trajectories are firstly obtained by tracking pixels with median filter, and five kinds of descriptors are extracted, namely trajectory shape, HOG, HOF, MBHx, and MBHy. iDTs improve the performance of DTs by taking into account camera motion correction. Generally speaking, iDTs resort to more sophisticated engineering skills and integrate much richer low-level visual cues compared with STIPs. Therefore, they represent two different kinds of low level features, namely sparse features and dense features, and may exhibit different properties with respect to variants of BoVWs.

\subsection{Feature Pre-processing}
The low-level local descriptors are usually high dimensional and strong correlated, which results in great challenges in the subsequent unsupervised learning such as $k$-means clustering and GMM training. Principal component analysis (PCA) \cite{Bishop06} is a statistical procedure to pre-process these features, which uses orthogonal transform to map feature into a set of linearly uncorrelated variables called principal components. Typically, the number of used principal components is less than the number of original variables, thus resulting in dimension reduction. Whitening technique usually follows the PCA, which aims to ensure the feature have the same variance through different dimensions. The transform formula of pre-processing is as followings:
\begin{equation}
  \mathbf{x} = \Lambda U^\top \mathbf{f},
\end{equation}
where $\mathbf{f}\in R^M$ is the original feature, $\mathbf{x} \in R^N$ is the PCA-whitened result, $U \in R^{M \times N}$ is the dimension reduction matrix from PCA, $\Lambda$ is the diagonal whitening matrix $diag(\Lambda) = [1/\sqrt{\lambda_1},\cdots,1/\sqrt{\lambda_N}]$, and $\lambda_i$ is the $i^{th}$ largest eigenvalue of covariance matrix.

It is worth noting that this step is not necessary and many previous encoding approaches skip this step, such as Vector Quantization \cite{SivicZ03}, Sparse Coding \cite{YangYGH09}, and Vector of Locally Aggregated Descriptor \cite{JegouPDSPS12}. However, in our evaluation, we found this step is of great importance to improve the recognition performance.

\subsection{Codebook Generation}
In this section, we present the codebook generation algorithms used for the following feature encoding methods. Generally there are two kinds of approaches: (i) partitioning the feature space into regions, each of which is represented by its center, called codeword, and (ii) using generative model to capture the probability distribution of features. $k$-mean \cite{Bishop06} is a typical method for the first type, and Gaussian Mixture Model (GMM) \cite{Bishop06} is widely used for the second.

\textbf{$k$-means.} There are many vector quantization methods  such as $k$-means clustering \cite{Bishop06}, hierarchical clustering \cite{Johnson67}, and spectral clustering \cite{NgJW01}. Among them, $k$-means is probably the most popular way to construct codebook. Given a set of local features $\{\mathbf{x}_1,\cdots,\mathbf{x}_M\}$, where $\mathbf{x}_m \in \mathbb{R}^D$. Our goal is to partition the feature set into $K$ clusters $\{\mathbf{d}_1,\cdots,\mathbf{d}_K\}$, where $\mathbf{d}_k \in \mathbb{R}^D$ is a prototype associated with the $k^{th}$ cluster. Suppose for each feature $\mathbf{x}_m$, we introduce a corresponding set of binary indicator variables $r_{mk} \in \{0,1\}$. If descriptor $\mathbf{x}_m$ is assigned to cluster $k$, then $r_{mk}=1$ and $r_{mj} = 0$ for $j \neq k$. We can then define an objective function:
\begin{equation}
  \min \mathcal{J}(\{r_{mk}, \mathbf{d}_k\}) = \sum_{m=1}^M \sum_{k=1}^K r_{mk} \| \mathbf{x}_m - \mathbf{d}_k \|_{2}^2.
  \label{equ:kmeans}
\end{equation}
The problem is to find values for $\{r_{mk}\}$ and $\{\mathbf{d}_k\}$ to minimize the objective function $\mathcal{J}$. Usually,  we can optimize it in an iterative procedure where each iteration involves two successive steps corresponding to optimization with respect to the $r_{nk}$ and $\mathbf{d}_k$. The details can be found in \cite{Bishop06}.

\textbf{GMM.} Gaussian Mixture Model is a generative model to describe the distribution over feature space:
\begin{equation}
  p(\mathbf{x}; \theta) = \sum_{k=1}^K \pi_k \mathcal{N}(\mathbf{x} ; \mu_k, \Sigma_{k}),
  \label{equ:gmm}
\end{equation}
where $K$ is mixture number, and $\theta=\{\pi_1,\mu_1,\Sigma_1,\cdots,$ $\pi_K,\mu_K,\Sigma_K\}$ are model parameters. $\mathcal{N}(\mathbf{x}; \mu_k, \Sigma_k)$ is $D$-dimensional Gaussian distribution.

Given the feature set $\mathbf{X} = \{\mathbf{x}_1,\cdots,\mathbf{x}_M\}$, the optimal parameters of GMM are learned through maximum likelihood estimation $\mathrm{arg}\max_{\theta} \ln p(\mathbf{X}; \theta)$. We use the iterative EM algorithm \cite{Bishop06} to solve this problem.

$k$-means algorithm performs a \emph{hard} assignment of feature descriptor to codeword, while the EM algorithm of GMM makes \emph{soft} assignment of feature to each mixture component based on  posterior probabilities $p(k|x)$. But unlike $k$-means, GMM delivers not only the mean information of code words, but also the shape of their distribution.

\subsection{Encoding Methods}
In this section, we provide a detailed description of thirteen feature encoding methods. According to the characteristics of encoding methods, they can be roughly classified into three groups, namely (i) voting based encoding method, (ii) reconstruction based encoding method, and (iv) super vector encoding method, as shown in Table \ref{tab:list}.

Let $\mathbf{X}$ be a set of $D$-dimensional local descriptors extracted from a video, $\mathbf{X}= [\mathbf{x}_1, \mathbf{x}_2,\cdots,\mathbf{x}_N] \in \mathbb{R}^{D \times N}$. Given a codebook with $K$ codewords, $\mathbf{D} = \left[\mathbf{d}_1, \mathbf{d}_2, \cdots, \mathbf{d}_K \right] \in \mathbb{R}^{D \times K}$. The objective of encoding is to compute a code $\mathbf{s}$ (or $\mathcal{S}$) \footnote{We use $\mathbf{s}$ to denote the code of voting and reconstruction based encoding methods, and $\mathcal{S}$ to represent the one of super vector based encoding methods.} for input $\mathbf{x}$ with $\mathbf{D}$. Table \ref{tab:list} lists all the formulations and dimension of encoding methods, where $\mathbf{s}(i)$ denotes the $i^{th}$ element of $\mathbf{s}$.

\subsubsection{Voting based encoding methods}
\begin{figure*}[htb]
  \centering
  \subfigure[VQ]{\includegraphics[width=0.24\linewidth]{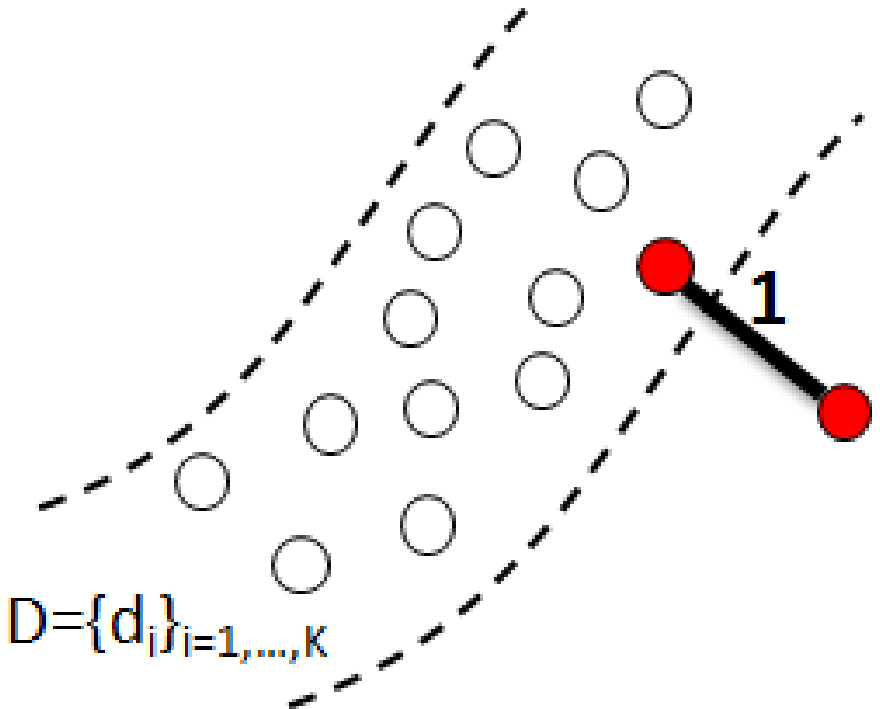}}
  \subfigure[SA]{\includegraphics[width=0.24\linewidth]{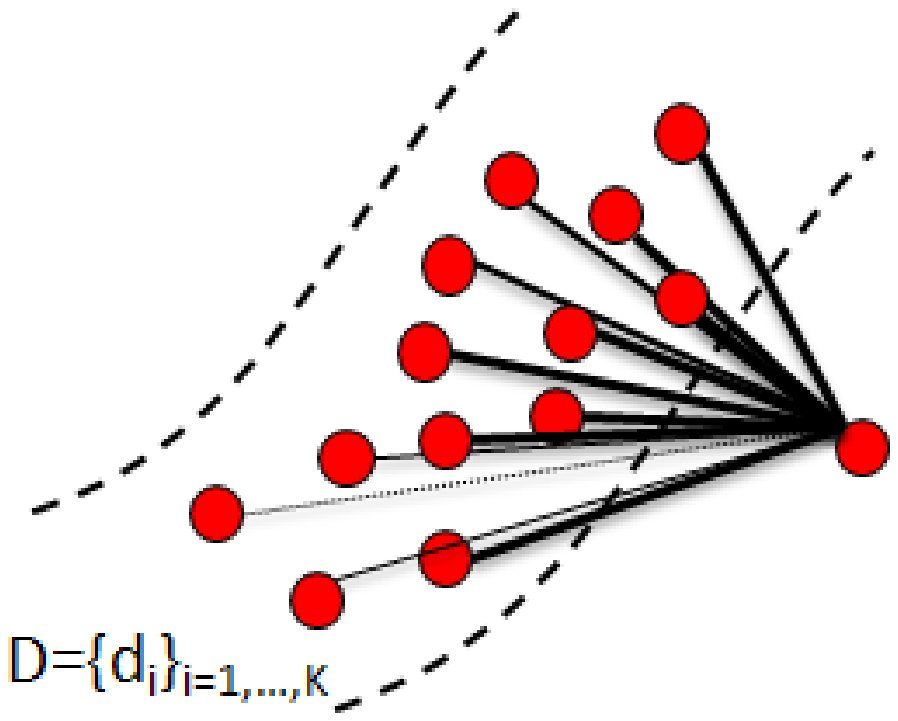}}
  \subfigure[SA-$k$]{\includegraphics[width=0.24\linewidth]{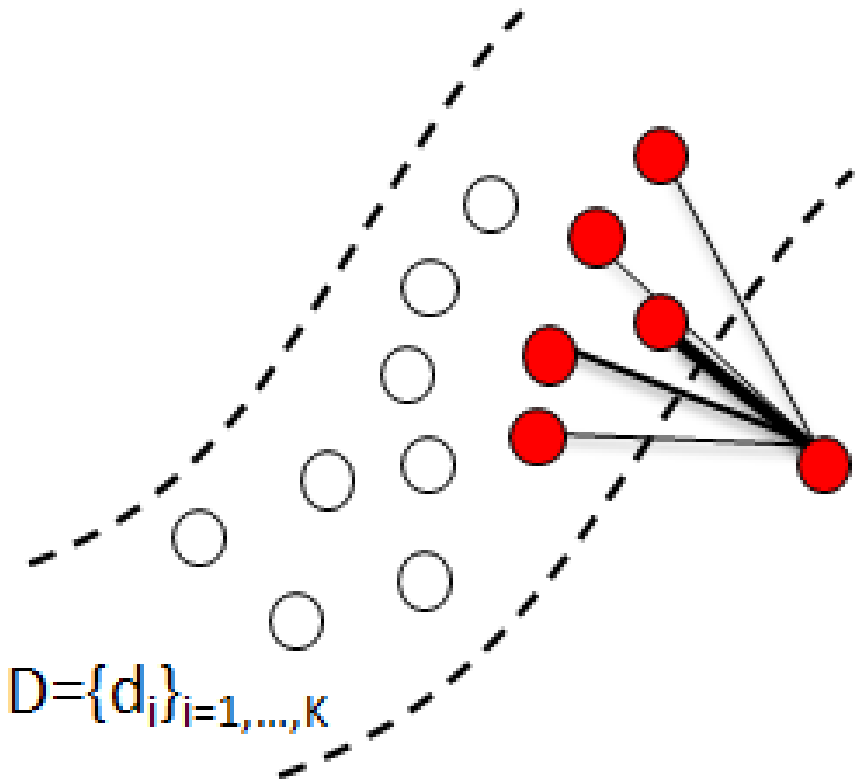}}
  \subfigure[SC/GSC]{\includegraphics[width=0.24\linewidth]{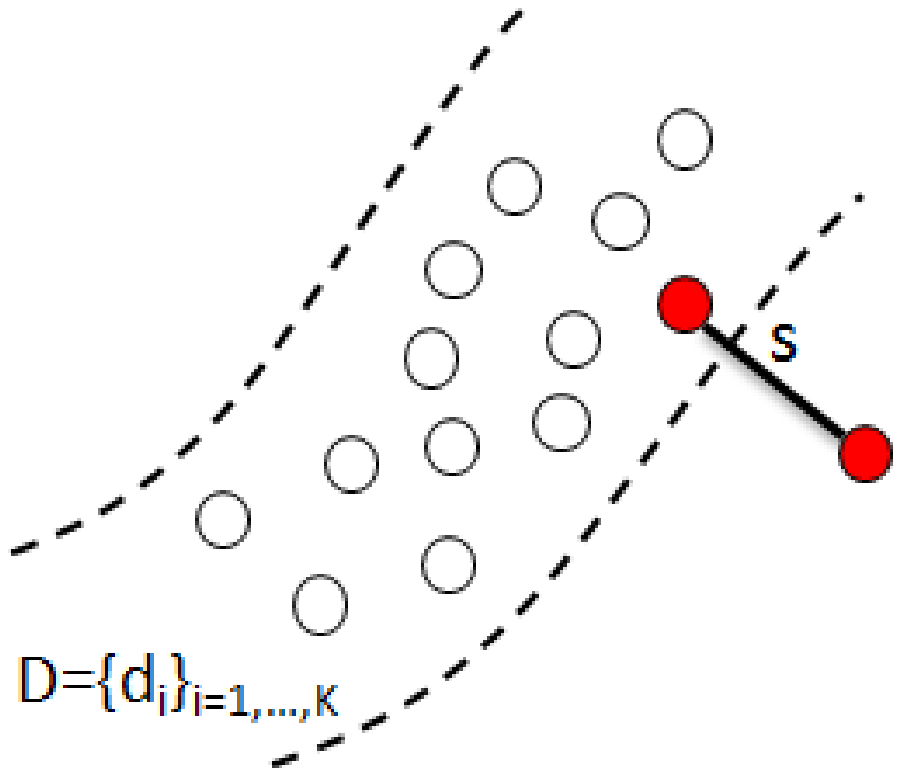}}
  \caption{Comparison among all the voting based encoding methods.}
  \label{fig:encVote}
\end{figure*}

Voting based encoding methods \cite{SivicZ03,GemertVSG10,LiuWL11,HuangHYT11,WuHWT12} are designed from the perspective of encoding process and each descriptor directly votes for the codeword using a specific strategy. A $K$-dimensional ($K$ is the size of codebook) code $\mathbf{s}$ is constructed for each single descriptor to represent the votes of the whole codebook. Methods along this line include Vector Quantization(or Hard Voting) \cite{SivicZ03}, Soft Assignment (or Kernel Codebook Coding) \cite{GemertVSG10}, Localized Soft Assignment \cite{LiuWL11}, Salient Coding \cite{HuangHYT11}, and Group Salient Coding \cite{WuHWT12}, as shown in Figure \ref{fig:encVote}.

For each descriptor $\mathbf{x}$, the voting value for the codeword $\mathbf{d}_i$ can be viewed as a function of $\mathbf{x}$, namely $\mathbf{s}(i)=\phi(\mathbf{x})$. Different encoding methods differ in the formulation of $\phi(\mathbf{x})$. For encoding of Vector Quantization (VQ):
\begin{equation}
\textrm{VQ:} \quad \phi(\mathbf{x}) = \left\{
 \begin{array}{l}
 1, \ \textrm{if} \ \ i=\textrm{arg}\min_{j}||\mathbf{x} - \mathbf{d}_j||_{2}, \\
 0,\ \ \textrm{otherwise,}
 \end{array} \right.
  \label{equ:vq}
\end{equation}
where each descriptor $\mathbf{x}$ only votes for its nearest codeword. The VQ encoding method can be viewed as a hard quantization and may cause much information loss. To encounter this problem, Soft Assignment (SA) encoding method votes for all the codewords:
\begin{equation}
\textrm{SA:} \quad \phi(\mathbf{x}) = \omega_i,
  \label{equ:sa}
\end{equation}
where $\omega_i$ is the normalized weight of descriptor $\mathbf{x}$ with respect to codeword $\mathbf{d}_i$:
\begin{equation}
\omega_i = \frac{\exp(-\beta\|\mathbf{x} - \mathbf{d}_i\|_2^2)}{\sum_{j=1}^K \exp(-\beta \|\mathbf{x} - \mathbf{d}_j\|_2^2)},
\label{equ:weight1}
\end{equation}
where $\beta$ is a smoothing factor controlling the softness of the assignment. Considering the manifold structure in the descriptor space, localized Soft Assignment (SA-$k$) votes for its $k$-nearest codewords:
\begin{equation}
\textrm{SA-}k: \quad \phi(\mathbf{x}) = \omega'_i = \frac{I(\mathbf{x},\mathbf{d}_i)\exp(-\beta\|\mathbf{x} - \mathbf{d}_i\|_2^2)}{\sum_{j=1}^K I(\mathbf{x},\mathbf{d}_j) \exp(-\beta \|\mathbf{x} - \mathbf{d}_j\|_2^2)},
\label{equ:weight2}
\end{equation}
where $I(\mathbf{x},\mathbf{d}_i)$ is the indicator function to identify whether $\mathbf{d}_i$ belongs to the $k$ nearest neighbor of $\mathbf{x}$:
\begin{equation}
    I(\mathbf{x},\mathbf{d}_i) = \left\{ \begin{array}{cl}
                                         1 & \ \ \mathrm{if} \ \ \mathbf{d}_i \in NN_k(\mathbf{x}), \\
                                         0 & \ \ \mathrm{otherwise.}
                                       \end{array}
                                       \right.
\end{equation}
Note that VQ can be viewed as a special case of SA-$k$ when $k$ is set as $1$.

Figure \ref{fig:encVote} illustrates the difference of these voting based encoding methods. VQ, Salient coding, and Group salient coding are all hard assignment strategies. Unlike VQ, the Salient coding employs the difference between the closest visual word and the other $k-1$ closest ones to obtain the voted weight but not 1. The detailed formulations of Salient coding and Group salient coding can be found in Table \ref{tab:list}.

\subsubsection{Reconstruction based encoding methods}
Reconstruction based encoding methods \cite{YangYGH09,WangYYLHG10,YuZG09,TroppG07} are designed from the perspective of decoding process, where the codes $\mathbf{s}$ are enforced to reconstruct the input descriptor $\mathbf{x}$. This kind of algorithm includes Orthogonal Matching Pursuit (OMP) \cite{TroppG07}, Sparse Coding (SPC) \cite{YangYGH09}, Local Coordinate Coding(LCC) \cite{YuZG09}, and Locality-constrained Linear Coding (LLC) \cite{WangYYLHG10}. Typically, these encoding methods are formulated in a least square framework with a regularization term:
\begin{equation}
\textrm{arg}\min_{\mathbf{s}}||\mathbf{x}-\mathbf{Ds}||^2_2 + \lambda\psi(\mathbf{s}),
\label{equ:reconEnc}
\end{equation}
where the least square term enforce the small reconstruction error, $\psi(\mathbf{s})$ encourages some properties of codes $\mathbf{s}$, $\lambda$ is a weight factor to balance this two terms.

Among these methods, OMP and SPC pursue a sparse representation. As for OMP, this constraint is conducted by $\ell_0$-norm:
\begin{equation}
\textrm{OMP:} \quad \psi(\mathbf{s})=||\mathbf{s}||_{0}
\end{equation}
where $\ell_0$-norm means the number of non-zero elements in $\mathbf{s}$. However, due to the non-convexity of $\ell_0$-norm, solution to this problem usually needs some heuristic strategy and obtains an approximate optimal solution. SPC relaxes this non-convex $\ell_0$-norm with $\ell_1$-norm:
\begin{equation}
\textrm{SPC:} \quad \psi(\mathbf{s})=||\mathbf{s}||_{1}
\end{equation}
where $\ell_1$-norm can also encourage the sparsity in code $\mathbf{s}$, and the solution is equal to the solution of $\ell_0$-norm under some conditions \cite{BrucksteinDE09}. The
$\ell_1$-norm relaxation allows for more efficient optimization algorithm \cite{LeeBRN06} and obtaining the global optimal solution.

OMP and SPC is empirically observed to tend to be local, \emph{i.e.} nonzero coefficients are often assigned to bases nearby to the encoded data \cite{YuZG09}. But this locality can not be ensured theoretically and they suggested a modification to SPC, called Local Coordinate Coding (LCC). This encoding method explicitly encourages the coding to be local, and they theoretically pointed out that under certain assumptions locality is more essential than sparsity, for successful nonlinear function learning using the obtained codes. Specifically, the LCC is defined as follows:
\begin{equation}
\textrm{LCC:} \qquad \psi(\mathbf{s}) = \| \mathbf{\hat e} \odot |\mathbf{s}| \|_{1}, \ \
\mathrm{s. t.} \ \ \mathbf{1}^T \mathbf{s} =1,
\label{equ:lcc}
\end{equation}
where $\odot$ denotes the element-wise multiplication, $\mathbf{\hat e}$ is the locality adaptor that give weights for each basis vector proportional to its similarity to the input descriptor $\mathbf{x}$:
\begin{equation}
\mathbf{\hat e} = \left[\mathrm{dist}(\mathbf{x},\mathbf{d}_1), \cdots, \mathrm{dist}(\mathbf{x},\mathbf{d}_K) \right]^\top,
\end{equation}
where $\mathrm{dist}(\mathbf{x},\mathbf{d}_k)$ is the Euclidean distance between $\mathbf{x}$ and $\mathbf{d}_k$. Due to the problem of $\ell_1$-norm optimization in both SPC and LCC, it is computationally expensive and hard to apply to large scale problem. Then, a practical coding scheme called Locality-constrained Linear Coding (LLC) \cite{WangYYLHG10} is designed, which can be viewed as a fast implementation of LCC that utilizes the locality constraint to project each descriptor into its local-coordinate system:
\begin{equation}
\textrm{LLC:} \qquad \psi(\mathbf{s}) = \| \mathbf{e} \odot \mathbf{s} \|^2_2, \ \
\mathrm{s. t.} \ \ \mathbf{1}^T \mathbf{s} =1,
  \label{equ:llc}
\end{equation}
where $\mathbf{e}$ is the exponentiation of $\mathbf{\hat e}$:
\begin{equation}
  \mathbf{e} =  \exp \left( \frac{\mathrm{dist}(\mathbf{x}, \mathbf{D})}{\sigma} \right),
\end{equation}
where $\sigma$ is used for adjusting the weighted decay speed for the locality adaptor. The constraint $\mathbf{1}^T \mathbf{s} =1$ follows the shift-invariant requirements of the final code vector. In practice, an approximate solution can be used to improve the computational efficiency of LLC. It directly selects the $k$ nearest basis vectors of $\mathbf{x}$ to minimize the first term in Equation (\ref{equ:reconEnc}) by solving a much smaller linear system. This gives the code coefficients for the selected $k$ basis vectors and other code coefficients are simply set to be zero.

\subsubsection{Super vector based encoding methods}
Super vector based encoding methods yield a very high dimensional representation by aggregating high order statistics. Typical methods include Local Tangent-based Coding (LTC) \cite{YuZ10}, Super Vector Coding (SVC) \cite{ZhouYZH10}, Vector of Locally Aggregated Descriptors (VLAD) \cite{JegouPDSPS12}, and Fisher Vector (FV) \cite{PerronninSM10} .

Local Tangent-based Coding \cite{YuZ10} assumes that codebook and descriptors are embedded in a smooth manifold. The main contents of LTC are manifold approximation and intrinsic dimensionality estimation. Under the Lipschitz smooth condition, the nonlinear function $f(\mathbf{x})$ can be approximated by a local linear function as:
\begin{equation}
  f(\mathbf{x}) \approx \sum^K_{i=1} \mathbf{s}(i)\left[f(\mathbf{d}_i)+0.5\nabla f(\mathbf{d}_i)^T(\mathbf{x} - \mathbf{d}_i)\right],
\end{equation}
where $\mathbf{s}(i)$ is obtained by LCC \cite{YuZG09}. Then, this approximate function can be viewed as a linear function of a coding vector $[\mathbf{s}(i),\mathbf{s}(i)(\mathbf{x}-\mathbf{d}_i)]_{i=1}^K \in \mathbb{R}^{K \times (1+D)}$. LTC argues that there is lower intrinsic dimensionality in the feature manifold. To obtain it, Principal Component Analysis (PCA) is applied to the term of $\mathbf{s}(i)(\mathbf{x}-\mathbf{d}_i)$ using a projection matrix $\mathbf{U}_i = [\mathbf{u}_1^i,\cdots, \mathbf{u}_C^i]\in \mathbb{R}^{D \times C}$ trained from training data, i.e., the local tangent directions of the manifold. Therefore, the final coding vector for LTC is written as follows:
\begin{equation}
\textrm{LTC:} \qquad  \mathcal{S} = \left[\alpha \mathbf{s}(i),\mathbf{s}(i)(\mathbf{x}-\mathbf{d}_i)^T\mathbf{U}_i \right]_{i=1}^K,
\end{equation}
where $\alpha$ is a positive scaling factor to balance the two types of codes. Super Vector Coding (SVC) \cite{ZhouYZH10} is a simple version of LTC. Unlike LTC, SVC yields the $\mathbf{s}(i)$ via VQ and does not apply PCA to the term of $\mathbf{s}(i)(\mathbf{x}-\mathbf{d}_i)$. Consequently, the coding vector of SVC is defined as follows:
\begin{equation}
  \textrm{SVC:} \ \mathcal{S} = [0,\mathbf{0},\cdots,\frac{\alpha\mathbf{s}(i)}{N\sqrt{p_i}},\frac{\mathbf{s}(i)}{N\sqrt{p_i}}(\mathbf{x}-\mathbf{d}_i),\cdots,0,\mathbf{0}],
\end{equation}
where $\mathbf{s}(i)=1$, $\mathbf{d}_i$ is the closest visual word to $\mathbf{x}$, and $\alpha$ is a positive constant.

Fisher vector is another super vector based encoding method derived from fisher kernel \cite{JaakkolaH98} and is introduced for large-scale image categorization \cite{PerronninSM10}. The fisher kernel is a generic framework which combines the benefits of generative and discriminative approaches. As it is known, the gradient of the log-likelihood with respect to a parameter can describe how that parameter contributes to the process of generating a particular example. Then the video can be described by the gradient vector of log likelihood with respect to the model parameters \cite{JaakkolaH98}:
\begin{equation}
  G_{\theta}^\mathbf{x} = \nabla_{\theta} \log p(\mathbf{x};\theta).
\end{equation}
Note that the dimensionality of this vector depends on the number of parameters in $\theta$. Perronnin \emph{et al}. \cite{PerronninSM10} developed an improved fisher vector which is as follows,
\begin{equation}
  \mathcal{G}_{\mu,k}^\mathbf{x} = \frac{1}{\sqrt{\pi_k}}  \gamma_k \left( \frac{\mathbf{x} - \mu_k}{\sigma_k}\right),
\end{equation}
\begin{equation}
  \mathcal{G}_{\sigma,k}^\mathbf{x} = \frac{1}{\sqrt{2\pi_k}} \gamma_k \left[ \frac{(\mathbf{x} - \mu_k)^2}{\sigma_k^2} - 1\right],
\end{equation}
where $\gamma_k$ is the weight of local descriptor $\mathbf{x}$ to $k^{th}$ Gaussian Mixture:
\begin{equation}
  \gamma_k = \frac{\pi_k \mathcal{N}(\mathbf{x}; \mu_k, \Sigma_k)}{\sum_{i=1}^K \pi_i \mathcal{N}(\mathbf{x}; \mu_i, \Sigma_i)}.
\end{equation}
The final fisher vector is the concatenation this two gradients:
\begin{equation}
  \mathrm{FV:} \qquad \mathcal{S} = [\mathcal{G}_{\mu,1}^{\mathbf{x}},\mathcal{G}_{\sigma,1}^\mathbf{x},\cdots,\mathcal{G}_{\mu,K}^{\mathbf{x}},\mathcal{G}_{\sigma,K}^\mathbf{x}].
\end{equation}
Vector of Locally Aggregated Descriptors (VLAD) \cite{JegouPDSPS12} can be viewed as a hard version of FV and only keeps the $1^{st}$ order statistics:
\begin{equation}
   \textrm{VLAD:} \qquad  \mathcal{S} = [\mathbf{0},\cdots,\mathbf{s}(i)(\mathbf{x}-\mathbf{d}_i),\cdots,\mathbf{0}],
\end{equation}
where $\mathbf{s}(i)=1$, $\mathbf{d}_i$ is the closest visual word to $\mathbf{x}$.

\subsubsection{Relations of Encoding Methods}
\label{sec:relation}
In this section, we summarize several practical techniques widely used in these encoding methods, and give a unified generative perspective of these encoding methods. This analysis will uncover the underline relations between these methods and provide insights for developing new encoding methods.

\textbf{From ``hard'' to ``soft''.} These encoding methods transform local features from descriptor space to codeword space. There are two typical transformation rules in these methods, namely \emph{hard assignment} and \emph{soft assignment}. Hard assignment quantizes the feature descriptor into a single codeword, while soft assignment enables the feature descriptor to vote for multiple codewords. In general, soft assignment accounts for the codeword uncertainty and plausibility \cite{GemertVSG10}, and reduces the information loss during encoding. This technical skill of soft assignment can be found in several encoding algorithms, such as SA-$all$ vs. VQ, and VLAD vs. Fisher Vector. By the same techniques, we can extend the VLAD to VLAD-$all$, SVC to SVC-$all$:
\begin{equation}
\mathrm{VLAD-}all: \ \mathcal{S} = [\omega_1(\mathbf{x}-\mathbf{d}_1),\cdots,\omega_K(\mathbf{x}-\mathbf{d}_K)],
\label{equ:vladall}
\end{equation}
\begin{equation}
\begin{split}
\mathrm{SVC-}all: \  \mathcal{S} & = \left[\frac{\alpha\omega_1}{N\sqrt{p_1}},\frac{\alpha\omega_1}{N\sqrt{p_1}}(\mathbf{x}-\mathbf{d}_1),\cdots, \right.  \\
& \left. \frac{\alpha\omega_K}{N\sqrt{p_K}},\frac{\alpha\omega_K}{N\sqrt{p_K}}(\mathbf{x}-\mathbf{d}_K) \right],
\end{split}
\label{equ:svcall}
\end{equation}
where $\omega_i$ is the normalized weight of feature descriptor $\mathbf{x}$ with respect to codeword $\mathbf{d}_i$ defined in Equation (\ref{equ:weight1}).

\textbf{From  ``global'' to ``local''.} In several encoding methods, the manifold structure in descriptor space is captured to improve the stability of encoding algorithms. In the traditional soft assignment, each descriptor is assigned with all the codewords, which is called \emph{global assignment}. However, in the high dimensional space of feature descriptor, Euclidian distance may be not reliable especially when the codeword is outside the neighborhood of feature descriptor. Therefore, in the encoding methods such as SA-$k$ and LLC, each descriptor is enforced to only vote for these codewords belonging to its $k$-nearest neighbors, called \emph{local assignment}. In general, the incorporation of local structure in encoding methods is able to improve the stability and reduce the sensitivity to noise in descriptor. Using the same techniques, we can also extend the VLAD-$all$ to VLAD-$k$, SVC-$all$ to SVC-$k$ by replacing the $\omega_i$ in Equation (\ref{equ:vladall}), (\ref{equ:svcall}) with localized $\omega_i'$ defined in Equation (\ref{equ:weight2}):
\begin{equation}
\mathrm{VLAD-}k: \qquad \mathcal{S} = [\omega'_1(\mathbf{x}-\mathbf{d}_1),\cdots,\omega'_K(\mathbf{x}-\mathbf{d}_K)],
\end{equation}
\begin{equation}
\begin{split}
\mathrm{SVC-}k: \qquad  \mathcal{S} & = \left[\frac{\alpha\omega'_1}{N\sqrt{p_1}},\frac{\alpha\omega'_1}{N\sqrt{p_1}}(\mathbf{x}-\mathbf{d}_1),\cdots, \right.  \\
& \left. \frac{\alpha\omega'_K}{N\sqrt{p_K}},\frac{\alpha\omega'_K}{N\sqrt{p_K}}(\mathbf{x}-\mathbf{d}_K) \right],
\end{split}
\end{equation}

\textbf{From ``zero order statistics'' to ``high order statistics''.}  In these super vector based encoding methods, they preserve not only the affiliations of descriptors to codewords (zero order statistics), but also the high order information such as the difference between descriptors mean and codeword, thus resulting a high-dimensional super vector representation. As these super vectors keep much richer information for each codeword, the codebook size is usually much smaller than that of voting and reconstruction based encoding methods. Above all, these super vector is with high dimension, storing more information, and is proved to outperform the other two kinds of encoding methods in Section \ref{sec:experiment}. The high dimensional super vector will be a promising representation and designing effective dimension reduction algorithms for super vector will be an interesting problem.

\textbf{Generative perspective of encoding methods.} Although these encoding methods are developed in different scenarios, a unified generative probabilistic model can be used to uncover the underline relations among them. These encoding methods can be interpreted in a latent generative model:
\begin{equation}
  \begin{split}
    p(\mathbf{h}) & \in \mathbb{P}, \\
    p(\mathbf{x}|\mathbf{h})&  = \mathcal{N}(\mathbf{x};W\mathbf{h} + \mu_\mathbf{x},\Sigma),
  \end{split}
\end{equation}
where $\mathbf{x} \in R^{D}$ represents the descriptor, $\mathbf{h} \in R^{K}$ denotes the latent factor, $\mathcal{N}(\mathbf{x};W\mathbf{h} + \mu_\mathbf{x},\Sigma)$ is multivariate Gaussian distribution. Different encoding methods mainly different in two aspects: \emph{How to model the prior distribution $\mathbb{P}$ of latent factor $\mathbf{h}$} and \emph{How to use the probabilistic model to transform the descriptor into codeword space.}

For encoding methods such as VQ, SA-$all$, VLAD-$all$, and Fisher vector, they choose the prior distribution $p(\mathbf{h})$ as follows:
\begin{equation}
  p(\mathbf{h}) = \prod_{i=1}^K \pi_i^{h_i},
\end{equation}
where $\mathbf{h}\in\{0,1\}^K$ is discrete random variable, and the prior distribution is a Multinomial distribution. For SA-$all$, this Mutinomial distribution is specified by uniform distribution, i.e. $\pi_1 = \cdots = \pi_K = \frac{1}{K}$, where for Fisher vector, this Multinomial distribution is learned during GMM training. Meanwhile the SA-$all$ choose the latent variable embedding to encodes the descriptor by computing conditional expectation, i.e. $\mathbf{s(x)} = \mathbb{E}(\mathbf{h}|\mathbf{x})$, while the Fisher vector choose the gradient embedding \cite{JaakkolaH98}, i.e.  $\mathcal{S}(\mathbf{x}) = \nabla_{\theta} \log p(\mathbf{x};\theta)$. The VQ encoding can be viewed an extreme case of Soft-$all$, when:
\begin{equation}
   p(\mathbf{x}|\mathbf{s})  = \mathcal{N}(\mathbf{x};W\mathbf{s} + \mu_\mathbf{x},\epsilon I), \ \ \epsilon \rightarrow 0.
\end{equation}
VLAD-$all$ and SVC-$all$ can be viewed as the gradient embedding in this extreme case.

For encoding methods such as sparse coding, the latent variable $\mathbf{h}$ is continuous and its corresponding prior distribution is specified as:
\begin{equation}
  p(\mathbf{h}) = \prod_{i=1}^K \frac{\lambda}{2} \exp(-\lambda|h_i|).
\end{equation}
This prior distribution is called Laplace prior and sparse coding can be viewed as the latent variable embedding of this generative model using the maximum a posteriori value (MAP), i.e. $\mathbf{s(x)} = \arg\max_{\mathbf{h}} p(\mathbf{h} | \mathbf{x})$.

\subsection{Pooling and Normalization Methods}
\begin{figure*}[!htb]
  \centering
  \includegraphics[width = \linewidth]{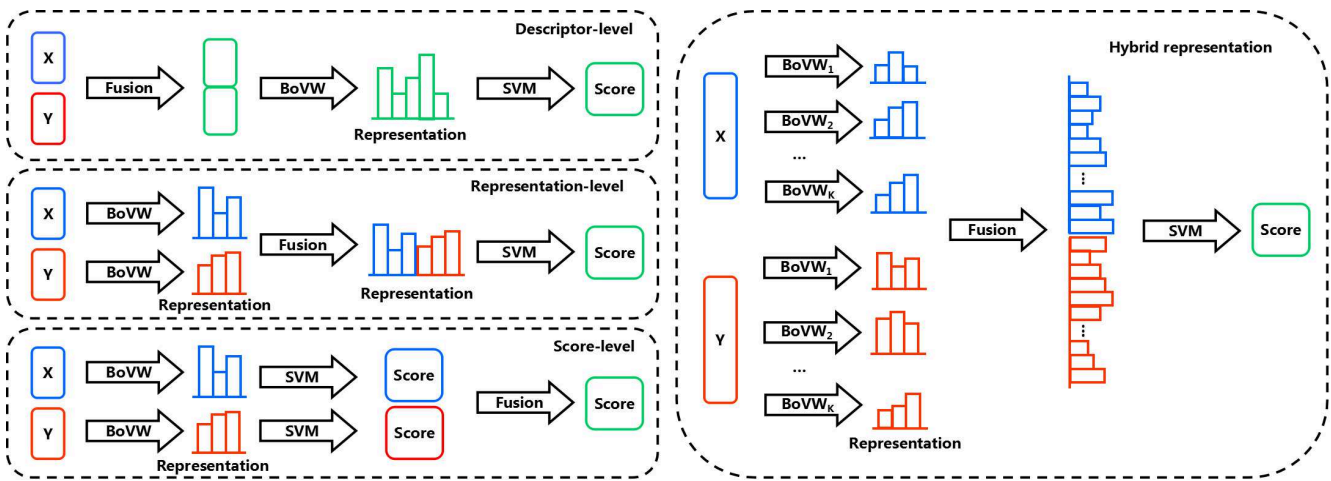}
  \caption{Feature fusion is performed in different levels: descriptor level, representation level, and score level. The complementary effect of varied BoVW models can also be taken into account and a hybrid representation is obtained by fusing outputs from different BoVW models.}
  \label{fig:fusion}
\end{figure*}

Given the code coefficients of all local descriptors in a video, a pooling operation is often used to obtain a global representation $\mathbf{p}$ for the video. Specifically, there are two common pooling strategies:
\begin{itemize}
\item \textbf{Sum Pooling.} With sum pooling scheme \cite{LazebnikSP06}, the $k^{th}$ component of $\mathbf{p}$ is $p_k = \sum_{n=1}^N \mathbf{s}_n(k)$.
\item \textbf{Max Pooling.} With max pooling scheme \cite{YangYGH09}, the $k^{th}$ component of $\mathbf{p}$ is $p_k = \max (\mathbf{s}_1(k),  \cdots, \mathbf{s}_N(k))$, where $N$ is the number of extracted local descriptors, $\mathbf{s}_n$ denotes the code of descriptor $\mathbf{x}_n$.
\end{itemize}
In \cite{BoureauPL10}, the authors presented a theoretical analysis of average pooling and max pooling. Their results indicate sparse features may prefer max pooling.\\

To make this representation invariant to the number of extracted local descriptors, the pooling result $\mathbf{p}$  is further normalized by some methods. Generally, there are three common normalization techniques:
\begin{itemize}
\item \textbf{$\ell_1$-Normalization.} In $\ell_1$ normalization \cite{YangYGH09}, the feature $\mathbf{p}$ is divided by its $\ell_1$-norm: $\mathbf{p} = \mathbf{p} / \| \mathbf{p}\|_{1}$.
\item \textbf{$\ell_2$-Normalization.} In $\ell_2$ normalization \cite{PerronninSM10}, the feature $\mathbf{p}$ is divided by its $\ell_2$-norm: $\mathbf{p} = \mathbf{p} / \| \mathbf{p}\|_{2}$.
\item \textbf{Power Normalization.} In power normalization \cite{PerronninSM10}, we apply in each dimension the following function:
\begin{displaymath}
  f (p_k) = \mathrm{sign}(p_k) | p_k |^\alpha.
\end{displaymath}
where $0 \leq \alpha \leq 1$ is a parameter for normalization. We can combine power normalization with $\ell_1$-normalization or  $\ell_2$-normalization.
\end{itemize}

Recently, a special normalization strategy is proposed for the VLAD, called \textbf{intra-normalization} \cite{ArandjelovicZ13}. In this paper, we extend it to all the super vector based encoding algorithms. This method carries out normalization operation in a block by block manner, where each block denotes the vector related to one codeword. Generally, the intra-normalization can be formulated as follows:
\begin{equation}
\mathbf{p} = \left[\frac{\mathbf{p}^1}{\| \mathbf{p}^1 \|},\cdots,\frac{\mathbf{p}^k}{\| \mathbf{p}^k \|},\cdots,\frac{\mathbf{p}^K}{\| \mathbf{p}^K \|}\right],
\end{equation}
where $\mathbf{p}^k$ denotes a vector related to codeword $\mathbf{d}_k$ (or the $k^{th}$ Gaussian), $\|\cdot\|$ may be $\ell_1$-norm or $\ell_2$-norm.

\section{Feature Fusion}
\label{sec:fusion}
Fusing multiple local features has turned out to be an effective method to boost the performance of recognition system in computer vision community \cite{GehlerN09,VedaldiGVZ09,TangYLK13,WangS13a,CaiWPQ14}. The video data is usually characterized in multiple views, such as static appearance, motion pattern, and motion boundary. The essence of multi-view data requires fusing different features for action recognition. In this section, we present several feature fusion methods for action recognition, and analyze its corresponding properties. Meanwhile, based on the analysis of fusion methods, we propose a simple yet effective representation, called \emph{hybrid representation}.

As shown in Figure \ref{fig:fusion}, the fusion methods are usually conducted in different levels, typically including: descriptor level, representation level, and score level. For descriptor level fusion, it is performed in the cuboid level, where multiple descriptors from the same cuboid are concatenated into a single one, and then it is fed into the BoVW to obtain the global representation. For representation-level fusion, it is performed in the video level, where different descriptors are input into BoVW separately and the resulting global representations are fused as a single one, which is further fed into classifier for recognition. For score-level fusion, it is also performed in the video level, but the representations of different descriptors are used independently for classifier training. The final recognition score is obtained by fusing the scores from multiple classifiers. For fusing the scores, arithmetical mean or geometrical mean is often used.

In general, these fusion methods at different levels owns their pros and cons, and the choice of fusion method should be guided by the dependence of descriptors. If these multiple descriptors from the same cuboid are highly correlated, it will be better to resort to descriptor level feature fusion. Otherwise, the choice of descriptor level fusion is not a good one, as descriptor level fusion usually results in a higher dimension and adds the difficulty for unsupervised feature learning such as $k$-means and sparse coding. For the case where different views of features are less correlated in cuboid level but highly correlated in video level, representation level fusion is usually a good choice. When these different features are independently with each other, it will be appropriate to choose score level fusion, as this fusion reduce the dimension for classifier training and make the learning faster and more stable.

The performance boosting of fusing multiple features mainly owns the complementarity of these features. However, the complementarity can be explored not only for different features, but also for different types of BoVW methods. As shown in Figure \ref{fig:fusion}, we propose a simple yet effective representation, called hybrid representation, which combines the outputs from multiple variants of BoVW and multiple descriptors. The resulting hybrid representation effectively explores the complementarity of different encoding methods and greatly enhances the descriptive power for action recognition. As we shall see in Section \ref{sec:stoa}, this representation will improve the recognition rate of a single BoVW model and obtain the state-of-the-art results on the three challenging datasets.

\section{Empirical Study}
\begin{figure*}[!htb]
  \centering
   \subfigure[HMDB51]{\includegraphics[width=0.48\linewidth]{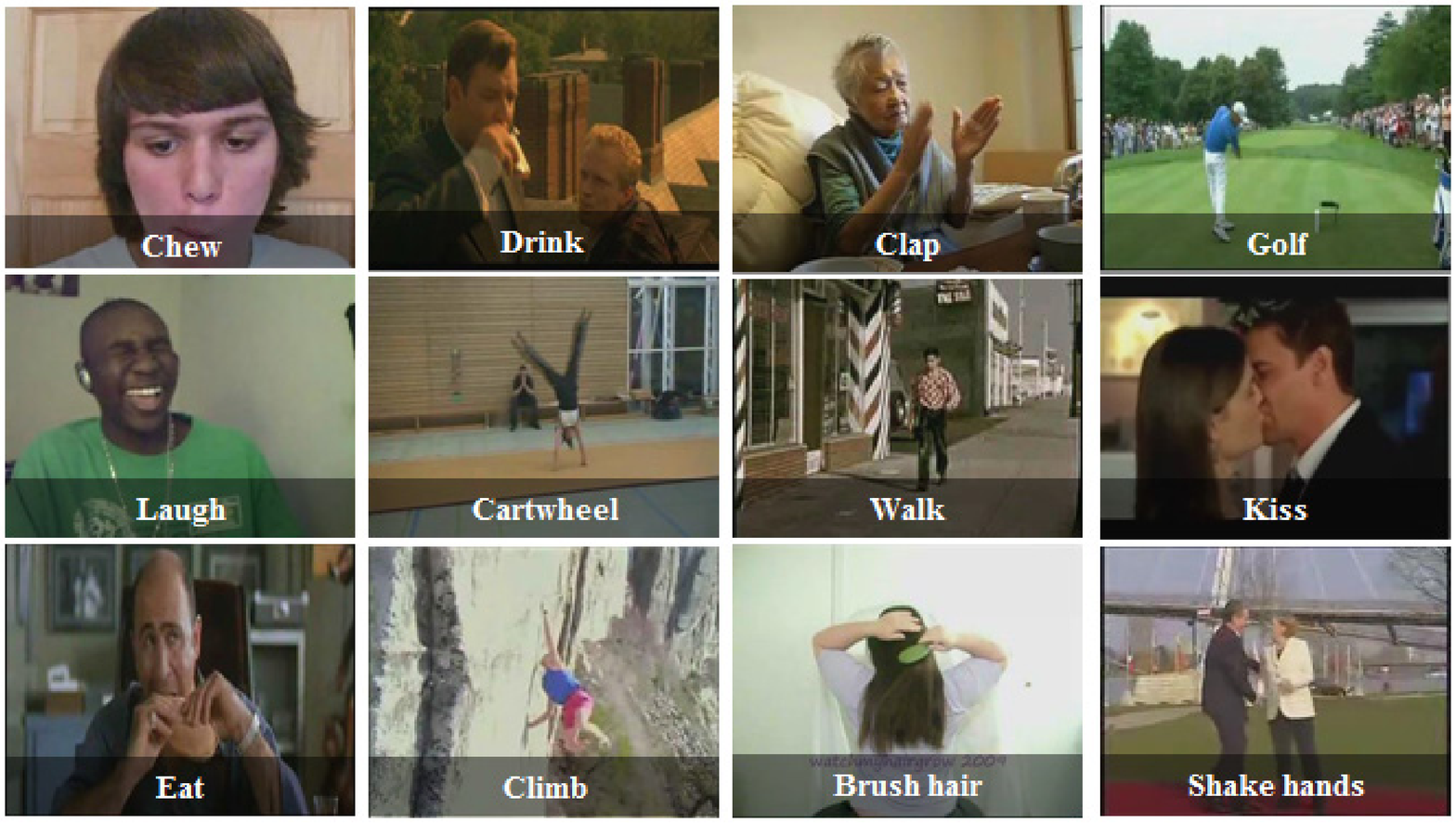}}
  \subfigure[UCF 50 and UCF101]{\includegraphics[width=0.48\linewidth]{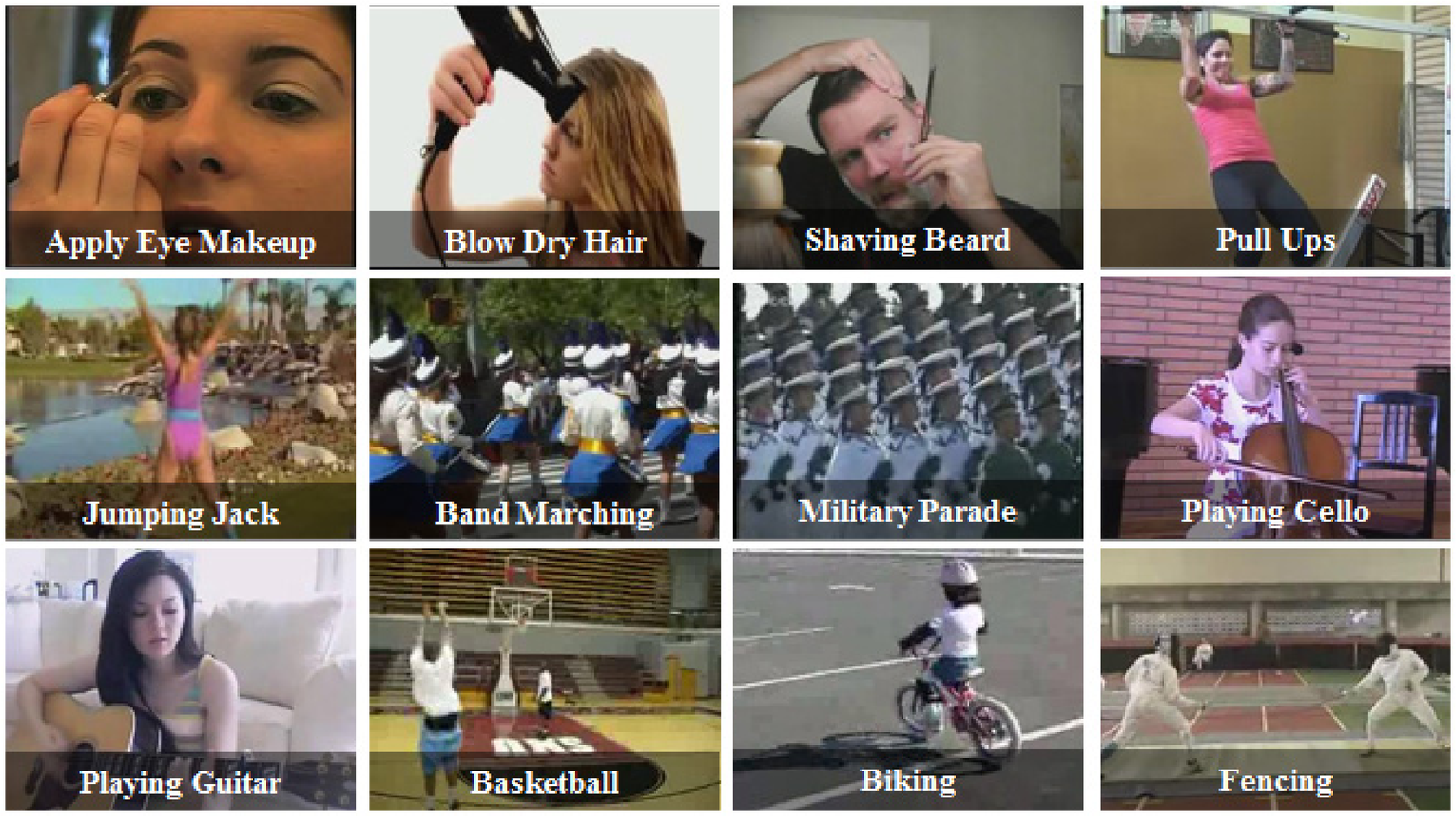}}
    \caption{Sample frames from the HMDB51, UCF50 and UCF101 datasets. Note that UCF50 is a subset of UCF101.}
  \label{fig:dataset}
\end{figure*}

\label{sec:experiment}
In this section, we describe the detailed experimental settings and the empirical study of variants of BoVW and different fusion methods. We first introduce the datasets used for evaluation and their corresponding experimental setup. We then extensively study different aspects of BoVW, including pre-processing techniques, encoding methods, pooling strategies, and normalization approaches. After that, we explore the different choices of fusion methods for multiple features. Finally, we compare the performance of our hybrid representation with that of the state-of-the-art methods on three challenging datasets.

\subsection{Datasets and Evaluation Protocols}
We conduct experiments on three public datasets: HMDB51 \cite{KuehneJGPS11}, UCF50 \cite{ReddyS13}, and UCF101 \cite{SOOMRO12}. Some examples of video frames are illustrated in Figure \ref{fig:dataset}. Totally, we work with 26,704 videos in this paper.

The \textbf{HMDB51 dataset} has 51 action classes with total 6,766 videos and each class has more than 100 videos \footnote{http://serre-lab.clps.brown.edu/resources/HMDB/index.htm}. All the videos are obtained from real world scenarios such as: movies, youtube. The intra-class variation is very high due to many factors, such as viewpoint, scale, background, illumination etc. Thus, HMDB51 is a very difficult benchmark for action recognition. There are three training and testing splits released on the website of this dataset. We conduct experiments based on these splits and report average accuracy for evaluation.

The \textbf{UCF50 dataset} has 50 action classes with total 6,618 videos, and each action class is divided into 25 groups with at least 100 videos for each class. The video clips in the same group are usually with similar background. We choose the suggested evaluation protocols of Leave One Group Out cross validation (LOGO) and report the average accuracy \cite{ReddyS13}.

The \textbf{UCF101 dataset} is an extension of the UCF50 dataset and has 101 action classes. The action classes can be divided into five types: human-object interaction, body-motion only, human-human interaction, playing musical instruments, and sports. Totally, it has 13,320 video clips, with fixed frame rate and resolution 25 FPS and 320 $\times$ 240 respectively. To our best knowledge, this dataset has been the largest dataset so far. We perform evaluation according to the three train/test splits released in Thumos'13 challenge \footnote{http://crcv.ucf.edu/ICCV13-Action-Workshop/} and report the mean average accuracy of these splits.

In our evaluation experiment, we choose linear Support Vector Machine (SVM) as our recognition classifier. Specifically, we use the implementation of LIBSVM \cite{ChangL11}. For multiclass classification, we adopt one-vs-all training scheme and choose the prediction with highest score as our predicted label.

\subsection{Local Features and Codebook Generation}
In our evaluation, we choose two widely-used local features, namely Space Time Interest Points (STIPs) \cite{Laptev05} with HOG, HOF descriptors \cite{LaptevMSR08}, and improved Dense Trajectories (iDTs) with HOG, HOF, MBHx, MBHy descriptors \cite{WangKSL13}. Specifically, we use the implementation released on the website of Laptev \footnote{http://www.di.ens.fr/~laptev/download.html} for STIPs and Wang \footnote{https://lear.inrialpes.fr/people/wang/improved\_trajectories} for iDTs. We choose the default parameter settings for both local features. STIPs and iDTs represent two types of local features: sparse interest points and densely-sampled trajectories. They may exhibit different properties with varying BoVW settings, and thus it is well worth exploring both STIPs and iDTs.

Regarding codebook generation, we randomly sample $100,000$ features to conduct $k$-means, where codebook size range from 1,000 to 10,000 for STIPs, and from 1,000 to 20,000 for iDTs. For GMM training, we randomly sample 256,000 features to learn GMMs with mixture number ranging from 16 to 512 for both STIPs and iDTs.

\subsection{Importance of Pre-processing}
\begin{figure}
  \centering
  \includegraphics[width=\linewidth,height=0.6\linewidth]{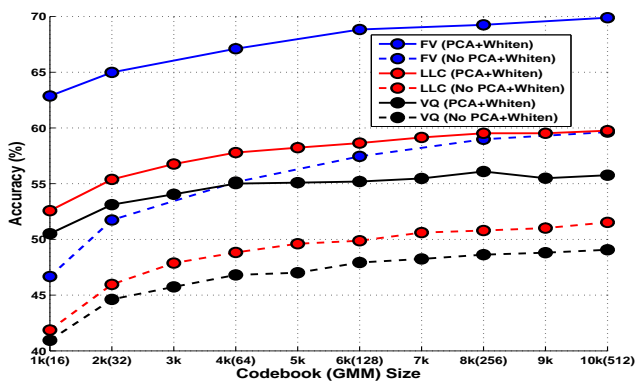}
  \caption{Comparison the results with PCA+Whiten and without PCA-Whiten of different encoding methods on the UCF101 dataset, where STIPs are chosen as the local features.}
  \label{fig:pca}
\end{figure}
In this section, we explore the importance of pre-processing in BoVW framework. Specifically, we use STIPs as local features and choose a representative method for each type of encoding, namely FV, LLC, and VQ . For pooling and normalization strategy, we use sum pooling and power $\ell_2$-normalization. We use the descriptor-level fusion method to combine HOG and HOF descriptors.

We conduct experiments on the UCF101 dataset and investigate the importance of pre-processing for these encoding methods. With pre-processing step, the descriptors of STIPs are firstly reduced to $100$-dimension and then whitened to have unit variance. The results are shown in Figure \ref{fig:pca}. We observe that the pre-processing technique of PCA-Whiten is very important to boost the performance of encoding methods. Surprisingly, the performance of FV (state-of-the-art) without PCA-Whiten is lower than or comparable to VQ and LLC with PCA-Whiten. In previous research work, PCA-Whiten is often done for FV encoding methods but seldom used for other encoding methods. Our study suggests that using PCA-Whiten techniques enable us to greatly improve final recognition rate for all encoding methods. We obtain the recognition rate $56.1\%$ for VQ, which significantly outperform over the result $43.9\%$ reported in \cite{SOOMRO12}, where the same local feature and encoding method is used.

In the remaining part of evaluation, we will use PCA-Whiten to de-correlate the descriptor, reduce the dimension, and normalize the variance. For descriptor level fusion of STIP, the dimension of concatenated descriptor is reduced from $162$ to $100$. For HOG and HOF, the dimension is reduced from $72$ to $40$, and from $90$ to $60$, respectively. For descriptor level fusion of iDT, the dimension of concatenated descriptor is reduced from $396$ to $200$. For separate descriptor, the dimensions of HOG, MBHx, and MBHy are all reduced from $96$ to $48$. HOF descriptor is reduced from $108$ to $54$.

\subsection{Exploration of Encoding Methods}
\label{sec:ex_encoding}
\begin{figure*}
  \centering
  \subfigure[HMDB51 with STIPs]{\includegraphics[width=0.49\linewidth,height=0.3\linewidth]{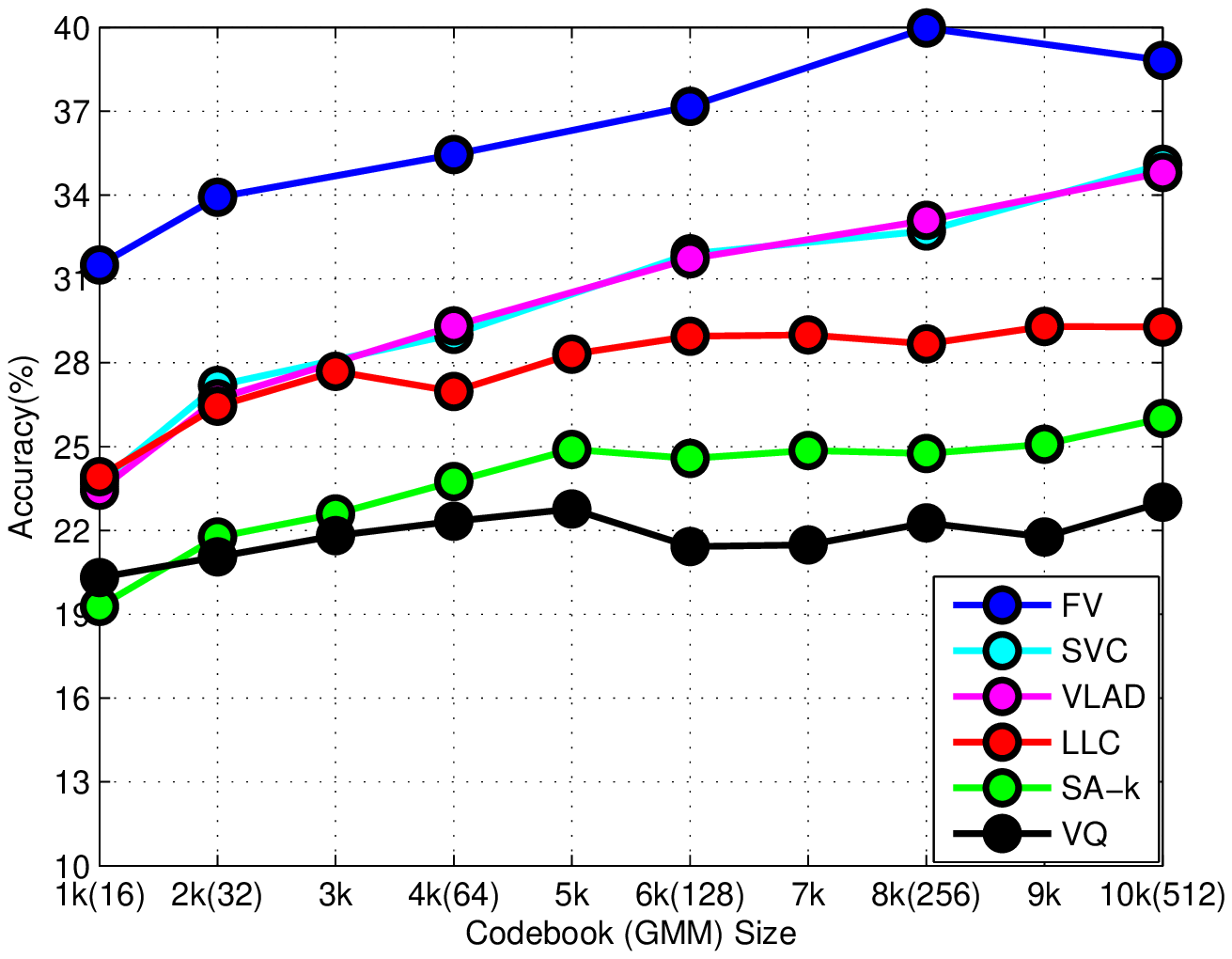}}
  \subfigure[HMDB51 with iDTs]{\includegraphics[width=0.49\linewidth,height=0.3\linewidth]{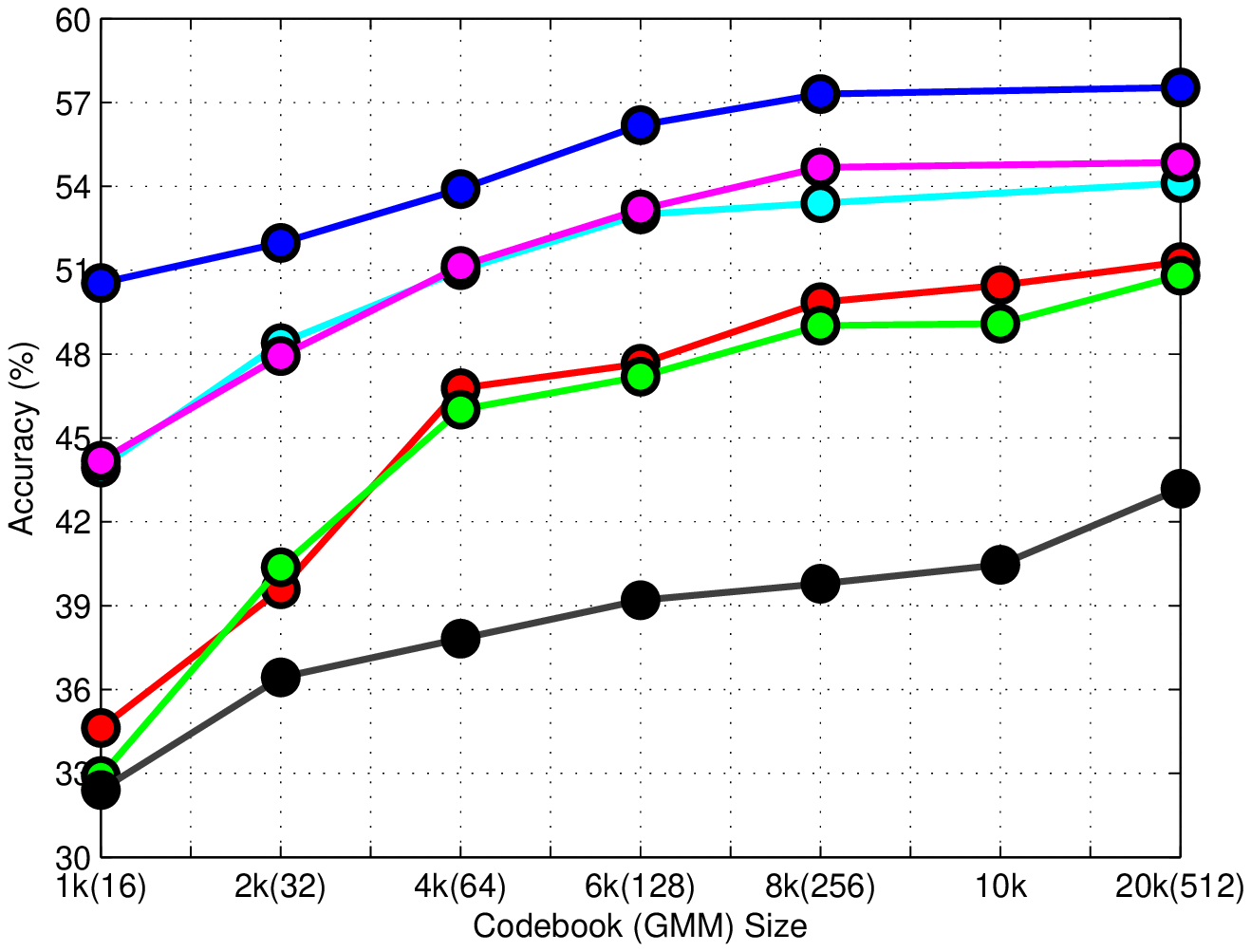}}\\
  \subfigure[UCF50 with STIPs]{\includegraphics[width=0.49\linewidth,height=0.3\linewidth]{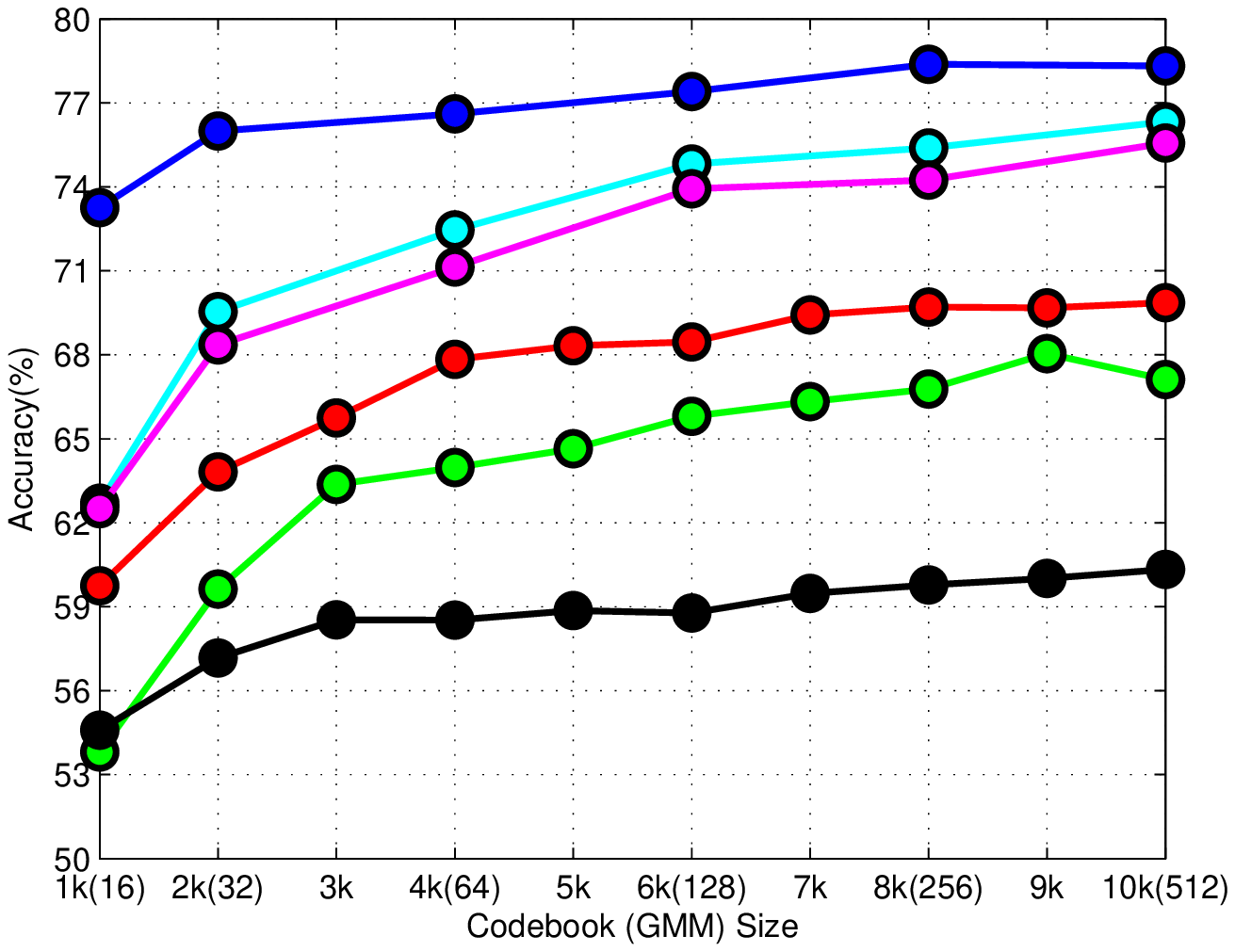}}
  \subfigure[UCF50 with iDTs]{\includegraphics[width=0.49\linewidth,height=0.3\linewidth]{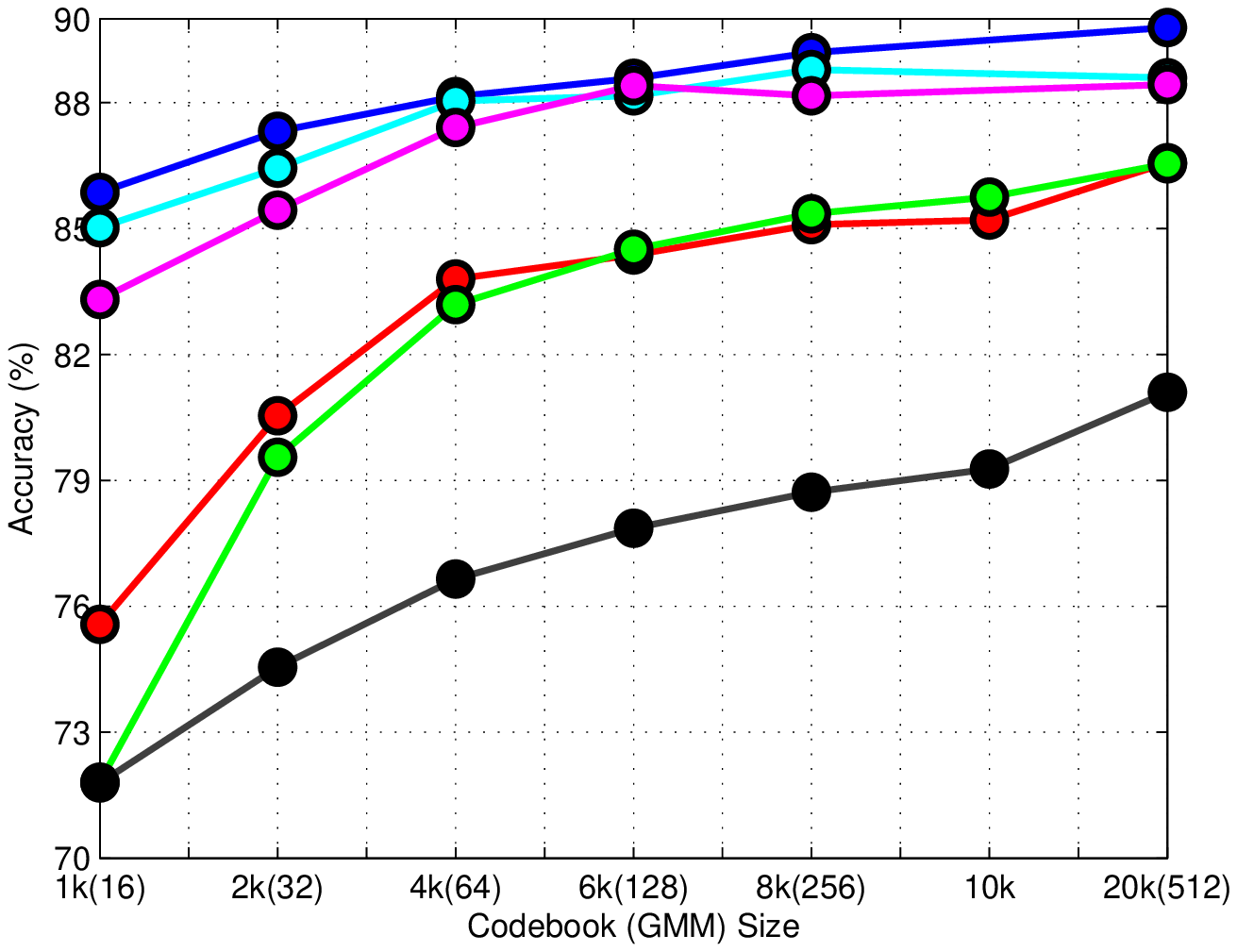}}\\
  \subfigure[UCF101 with STIPs]{\includegraphics[width=0.49\linewidth,height=0.3\linewidth]{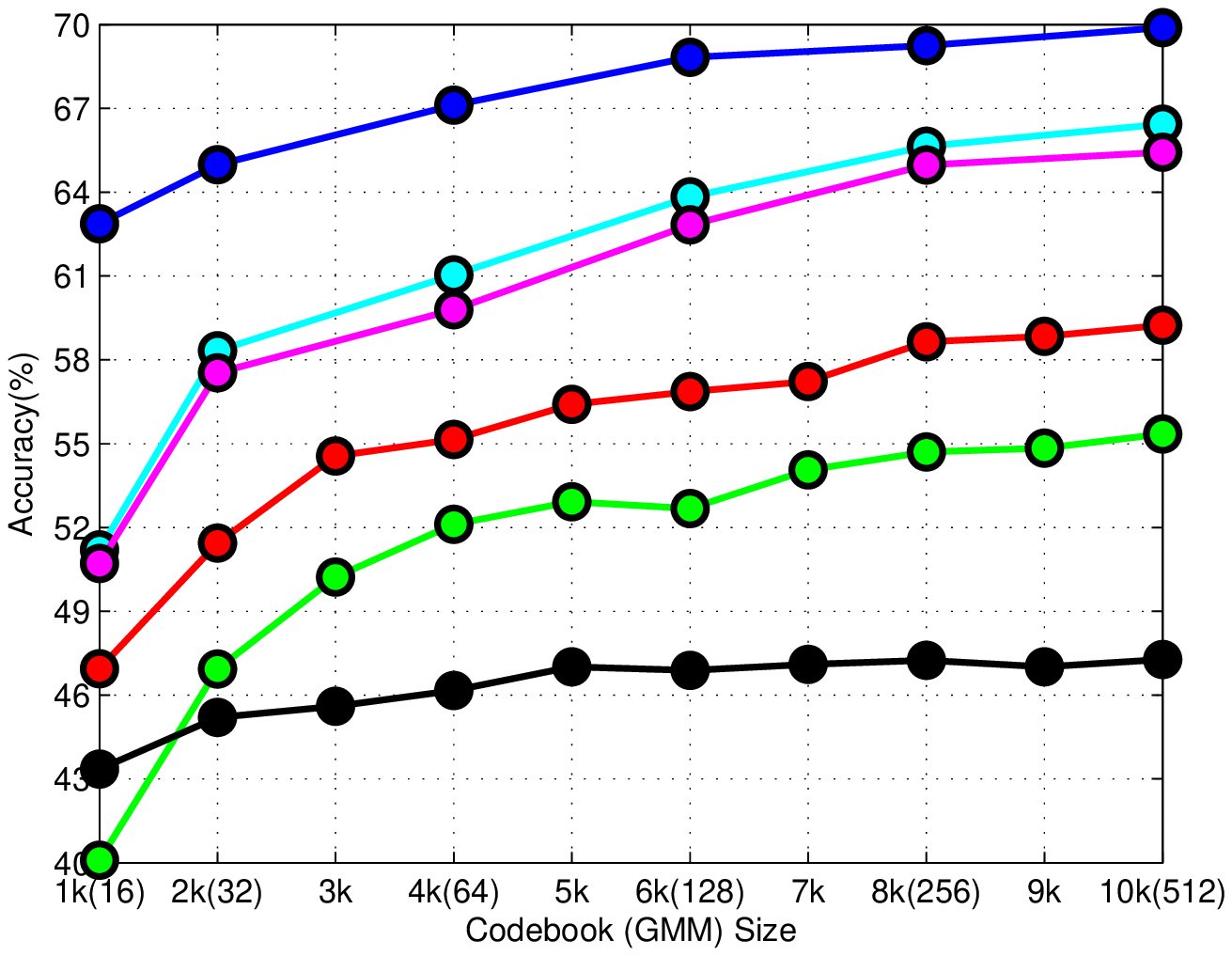}}
  \subfigure[UCF101 with iDTs]{\includegraphics[width=0.49\linewidth,height=0.3\linewidth]{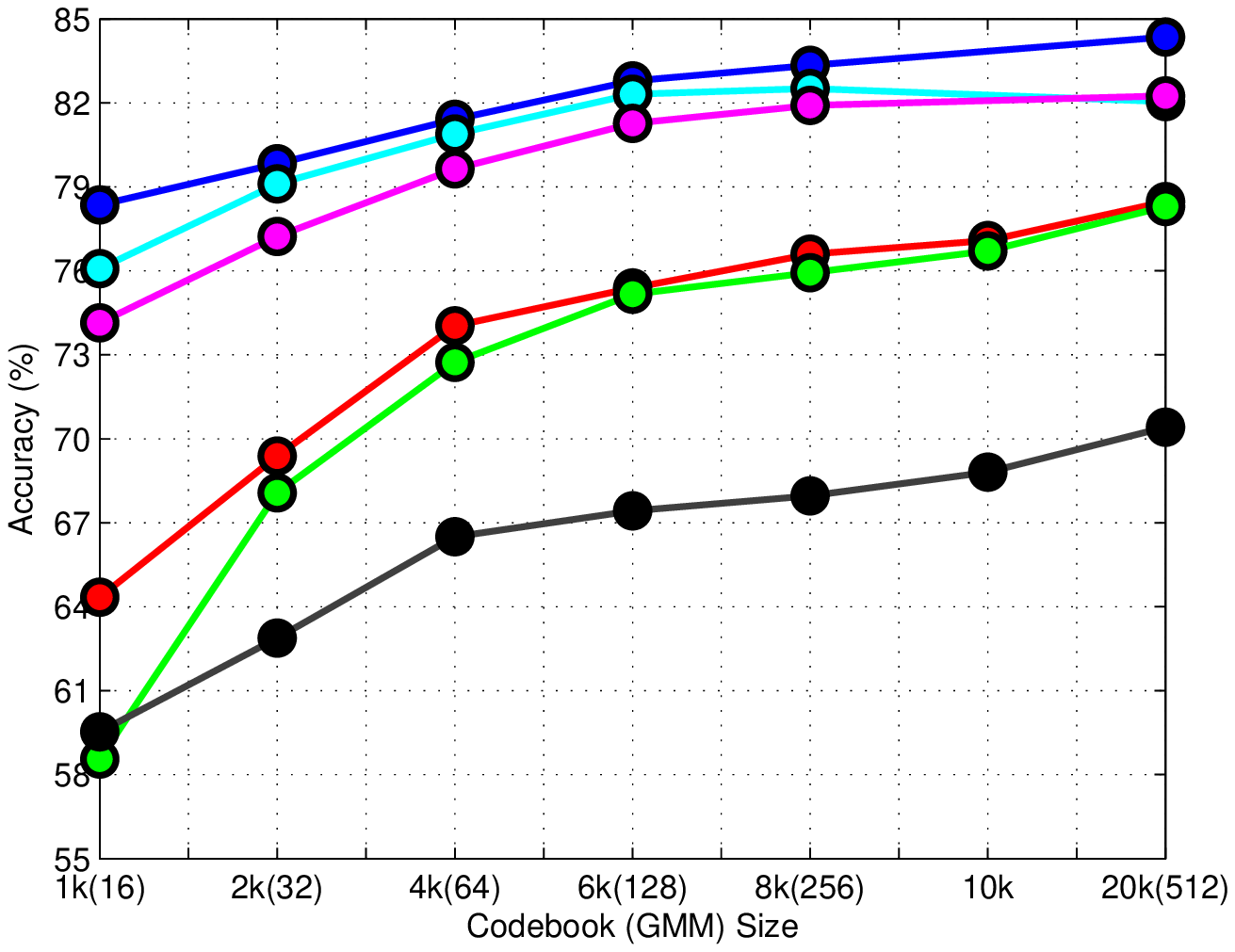}}
  \caption{Performance of different encoding methods with varying codebook (GMM) sizes on the HMDB51, UCF50, and UCF101 datasets for STIPs and iDTs features using descriptor-level fusion.}
  \label{fig:encIDT}
\end{figure*}
\begin{figure*}
  \centering
  \subfigure[Time of STIPs]{\includegraphics[width=0.49\linewidth,height=0.3\linewidth]{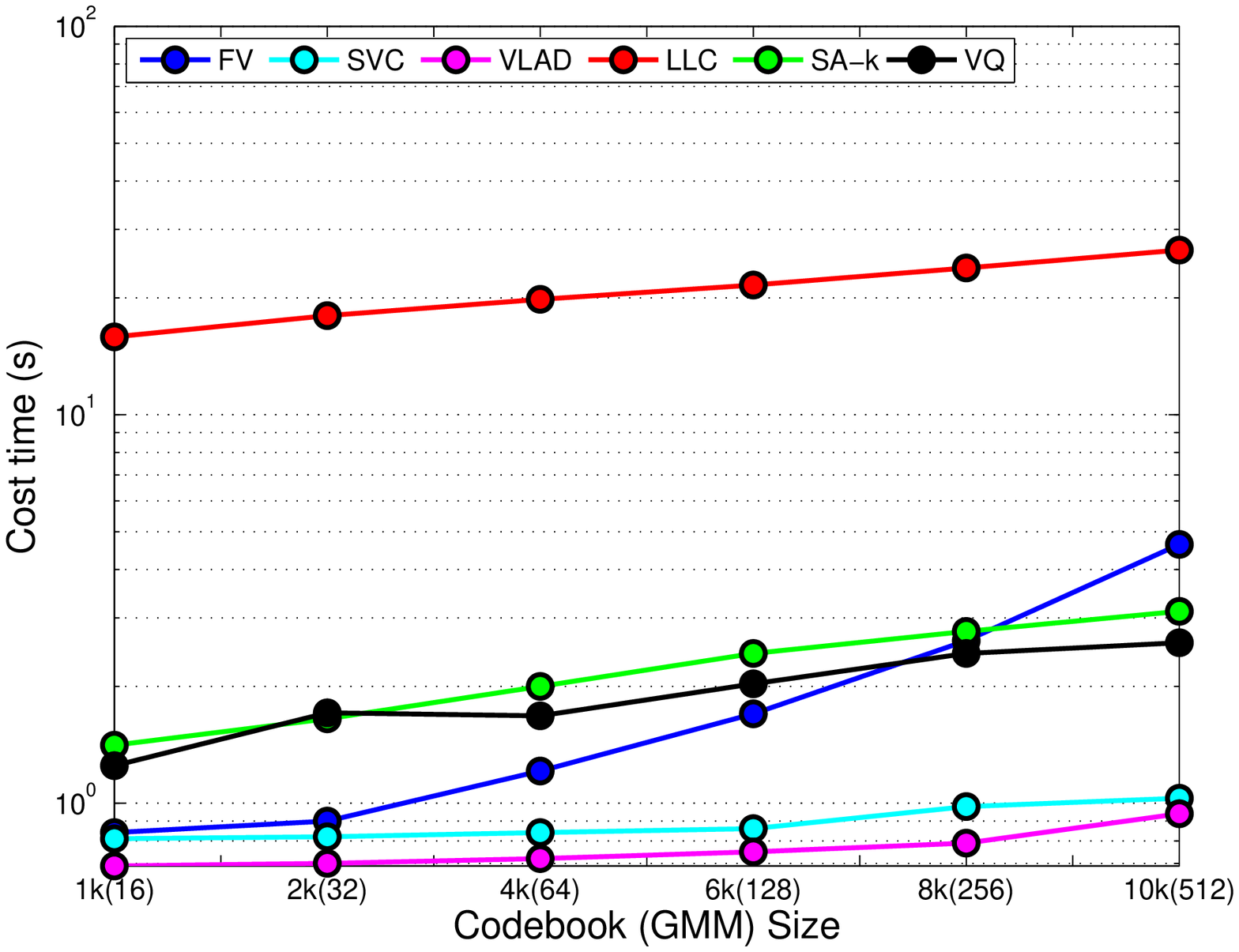}}
  \subfigure[Time of iDTs]{\includegraphics[width=0.49\linewidth,height=0.3\linewidth]{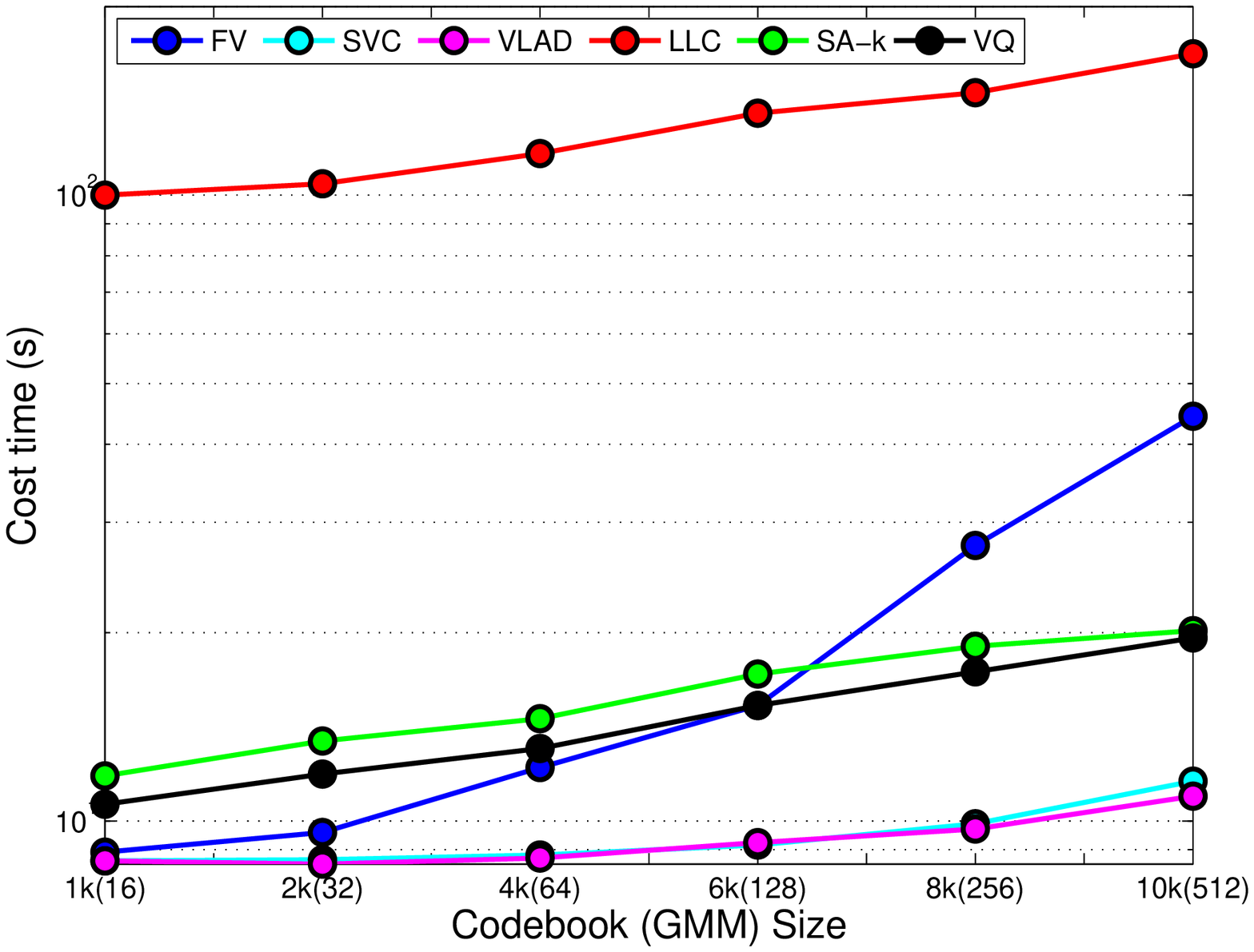}}\\
  \caption{Average time of different encoding methods with varying codebook (GMM) sizes on the UCF101 datasets for STIPs and iDTs features using descriptor-level fusion.}
  \label{fig:time}
\end{figure*}
In this section, we compare and analyze the performance of different encoding methods. For each encoding method, we fix other settings, such as parameter setting, pooling and normalization strategy, the same with previous papers. We explore these encoding methods with descriptor level fusion, for both STIPs and iDTs. The influence of different pooling and normalization strategy, and fusion methods will be investigated in the following sections.

\textbf{Encoding methods selection and setting}. We select six popular encoding methods according to categorization in Table \ref{tab:list}. For voting based encoding methods, we choose VQ as a baseline and SA-$k$ as a representative method. LLC is selected as the representative of reconstruction-based encoding methods due to its computational efficiency and performance \cite{WangWQ12}. Super vector based encoding methods have shown the state-of-the-art performance on several datasets \cite{WangS13a} . We choose three super vector based encoding methods for evaluation, namely FV, VLAD, and SVC.

\emph{Baseline: Vector Quantization Encoding (VQ).} In the baseline method, each descriptor is quantized into a single codeword. Following the suggested settings in object recognition \cite{ZhangMLS07}, the final histogram is obtained by sum pooling and normalized with $\ell_1$ norm.

\emph{Localized Soft Assignment Encoding (SA-$k$).} In the localized soft assignment, each descriptor is assigned to its corresponding $k$ nearest neighborhood. It requires a single parameter $\beta$, which is the smoothing factor controlling the softness. According to \cite{LiuWL11}, we set $\beta$ as $1$ and $k$ as 5 in our evaluation. We use max pooling and $\ell_2$-normalization.

\emph{Locality-constrained Linear Encoding (LLC).} Following \cite{WangYYLHG10}, we use approximated LLC for fast encoding, where we simply use $k$ nearest neighborhood of descriptor as the local bases. The parameter of $k$ is set as $5$, and we choose max pooling and $\ell_2$-normalization strategy.

\emph{Fisher Vector (FV).} For GMM training, we use the $k$-means result to initialize iteration and the covariance matrix of each mixture is set as a diagonal one. Following \cite{PerronninSM10}, we use sum pooling and power $\ell_2$-normalization.

\emph{Vector of Locally Aggregated Vector (VLAD).} VLAD was originally designed for image retrieval in \cite{JegouPDSPS12} and can be viewed as a simplified version FV for fast implementation. Just like FV, we choose sum pooling and power $\ell_2$-normalization.

\emph{Super Vector Coding (SVC).} From the view of statistics, SVC can be viewed as a combination of VQ and VLAD. It contains the zeros and first-order statistics, and the parameter $\alpha$ keep balance between these two components. Following \cite{ZhouYZH10}, we set $\alpha$ as $0.1$. Like other super vector based encoding methods, we choose sum pooling and power $\ell_2$-normalization.

\textbf{Results and analysis.} The experimental results of STIPs and iDTs on the three datasets are shown in Figure \ref{fig:encIDT}. Several rules can be found from these experimental results:
\begin{itemize}
  \item Basically, the recognition performance of all selected encoding methods increases as the size of codebook (GMM) becomes larger and will reach a plateau when the size exceeds a threshold. For super vector based encoding methods, the performances reach a saturation when size of codebook (GMM) becomes $256$ for both STIPs and iDTs. There is a slight change of the recognition rate when GMM size grows from $256$ to $512$. For the other two types of encoding methods, the performances are saturated as the size of codebook reaches $8,000$. We also notice that these encoding methods using iDTs have slight improvements when the codebook size varies from 8,000 to 20,000 , while the performances using STIPs start shaking when the codebook size becomes larger than 8,000 due to the over-fitting effect. This difference may be ascribed to the dimension of local descriptors and sampling strategy. The descriptors dimension of iDTs is twice of STIPs and requires more codewords to divide the feature space. Meanwhile, STIPs is a set of interest points and the extracted descriptors distribute sparsely in the feature space. The codebook with large size will result in an over-partition of feature space, which means for a specific video, there may be no descriptors falling into the corresponding regions for some codewords. iDTs are more densely sampled features and codebook with large size is more suitable to divide the space of dense features. \textbf{Above all, for a good balance between performance and efficiency, sizes of $256$ and $8,000$ are good choices for super vector based encoding and other encoding respectively}.
  \item For local features of both SITPs and iDTs, super vector based encoding methods outperform the other types of encoding methods on the three datasets. According to previous introduction, these super vector encoding methods not only preserve the affiliations of descriptors to codewords, but also keep high order information such as the difference of means and variances. These high order information enables the encoding methods to better capture the distribution shape of descriptor in feature space. In these super vector based methods, FV is typically better than VLAD and SVC, whose performance is quite similar. This can be own to two facts: (i) FV keeps both $1^{st}$ and $2^{nd}$ statistics, which is more informative than VLAD (only $1^{st}$ statistics) and SVC ($0^{th}$ statistics and $1^{st}$ statistics). (ii) FV is based on GMM and each descriptor is softly assigned to codewords using posterior probability, while VLAD and SVC are based on $k$-means results and use hard assignment. We also notice that the difference between FV and the other two methods (VLAD, SVC) for iDTs seems smaller than STIPs. The more dense descriptors may make the learned codebook more stable for SVC and VLAD, and reduce the influence of soft assignment in FV. Meanwhile, the information contained in $2^{nd}$ statistics may be less complementary to $1^{st}$ statistics for iDTs. \textbf{In conclusion, super vector based representation, aggregating high order information, is a more suitable choice for good performance, when the high dimension of representation is acceptable}.
  \item For reconstruction based and voting based encoding methods, VQ reaches the lowest recognition rate for STIPs and iDTs on the three datasets. This can be ascribed to the hard assignment and descriptor ambiguity in the VQ method. In essence, the LLC and SA-$k$ are quite similar in spirt, for that they both consider locality when mapping descriptor into codeword. The performance of LLC is better than SA-$k$ for STIPs, while the performances of them are almost the same for iDTs. This can be explained by the mapping strategy in LLC and SA-$k$. The mappings of descriptor to the nearest codewords in LLC are determined jointly according to their effect in minimizing the reconstruction error, while the mappings in SA-$k$ are calculated independently for each individual codeword according to the Euclidean distance. The mapping method in LLC may be more effective to deal with manifold structure than just considering Euclidean distance in SA-$k$. For sparse features such as STIPs, the descriptors distribute sparsely around each codeword, and using Euclidean distance may introduce noise and instability for SA-$k$. For dense features such as iDTs, the descriptors are usually sampled densely and more compact around codewords, reducing the influence caused by the usage of Euclidean distance. \textbf{In a word, compared with hard assignment, locality and soft assignment is an effective strategy to improve the performance of encoding methods}.
  \item STIPs and iDTs represents two types of local features, namely sparsely-sampled and densely-sampled features. In general, they exhibit consistent performance trends for different encoding methods, for example, super vector encoding methods outperforms others, soft-assignment is better than hard-assignment. However, there is a slight difference between them in some aspects, such as sensitivity to codebook size and encoding methods, performance gaps among super vector based methods, difference between LLC and SA-$k$, as previously observed. From the perspective of data manifold, the more densely-sampled features can help us more accurately describe the data structure in the feature space. We can obtain a more compact clustering result using $k$-means, and the local Euclidean distance is more stable. \textbf{Thus, when choosing codebook size and encoding method, the type of local feature can be a factor needed to be considered}.
\end{itemize}

\textbf{Computational costs.} We also compare the efficiency of different encoding methods and the running time is shown in Figure \ref{fig:time}. Our codes are all implemented in Matlab, and running on a workstation with 2x Intel Xeon 5560 2.8GHz CPU and 32G RAM. We randomly sample 50 videos from the UCF101 dataset and report the total time for these videos. For super vector based methods, FV is much slower due to the calculation of posterior probability during encoding, and the time of VLAD and SVC is almost the same. For the other types of encoding methods, LLC is less efficient as it solves a least square problem. The computational cost of super vector encoding methods are usually lower than that of the other types of encoding methods, due to their smaller codebook sizes.

Based on the above analysis, super vector based encoding methods are more promising for high performance and fast implementation, especially for SVC, VLAD. However, the feature dimension of super vector methods is much higher than the other two kinds of encoding methods, for example, when the codebook size is 256, the dimension of FV and VLAD is 102,400 and 51,200 respectively for iDT features. The effective dimension reduction may be a future research direction for super vector encoding methods.

\subsection{Exploration of Pooling and Normalization}
\label{sec:poolingnorm}
\begin{figure*}
  \center
\includegraphics[width=0.3\linewidth,height=0.3\linewidth]{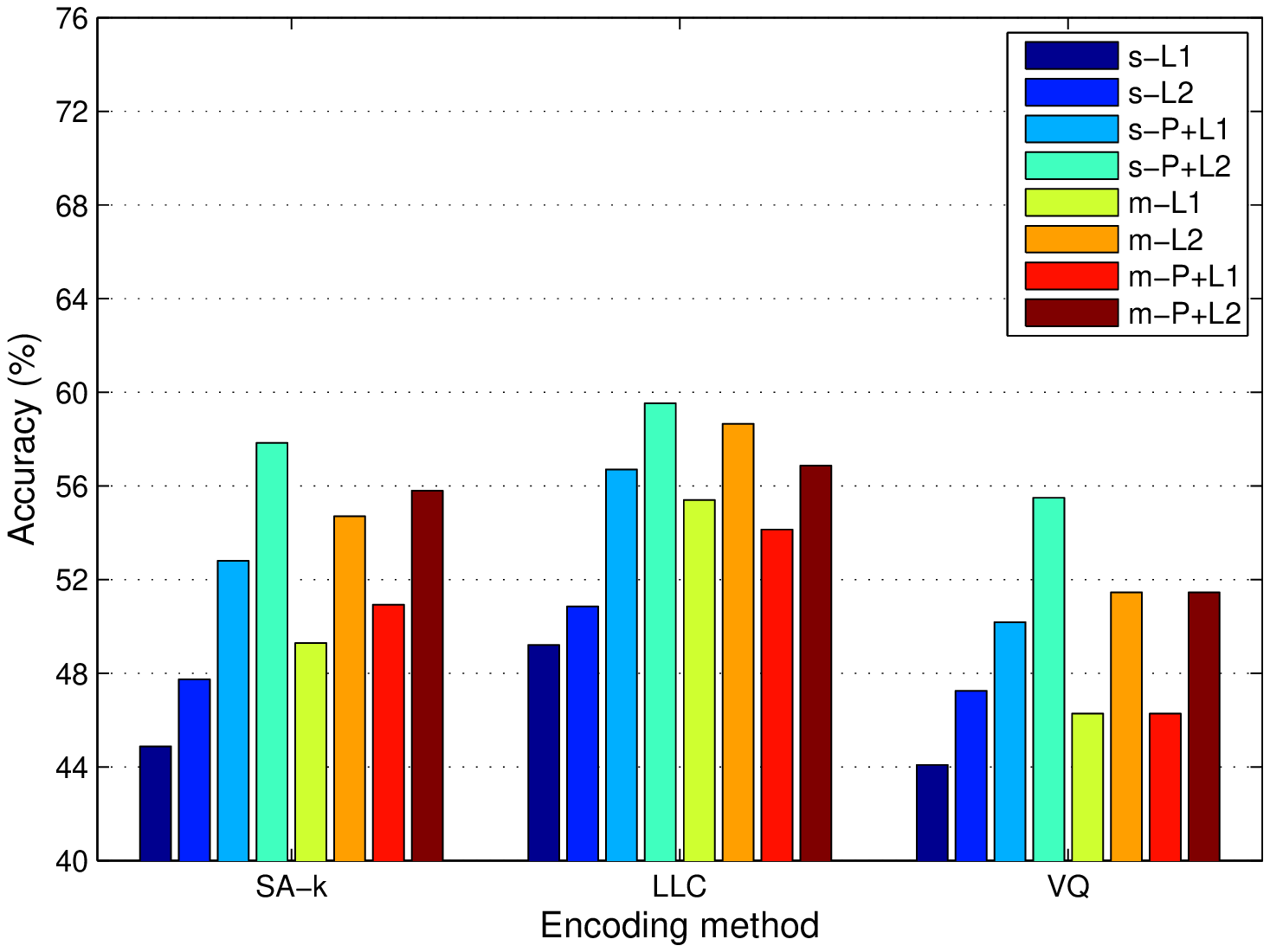}
\includegraphics[width=0.68\linewidth,height=0.3\linewidth]{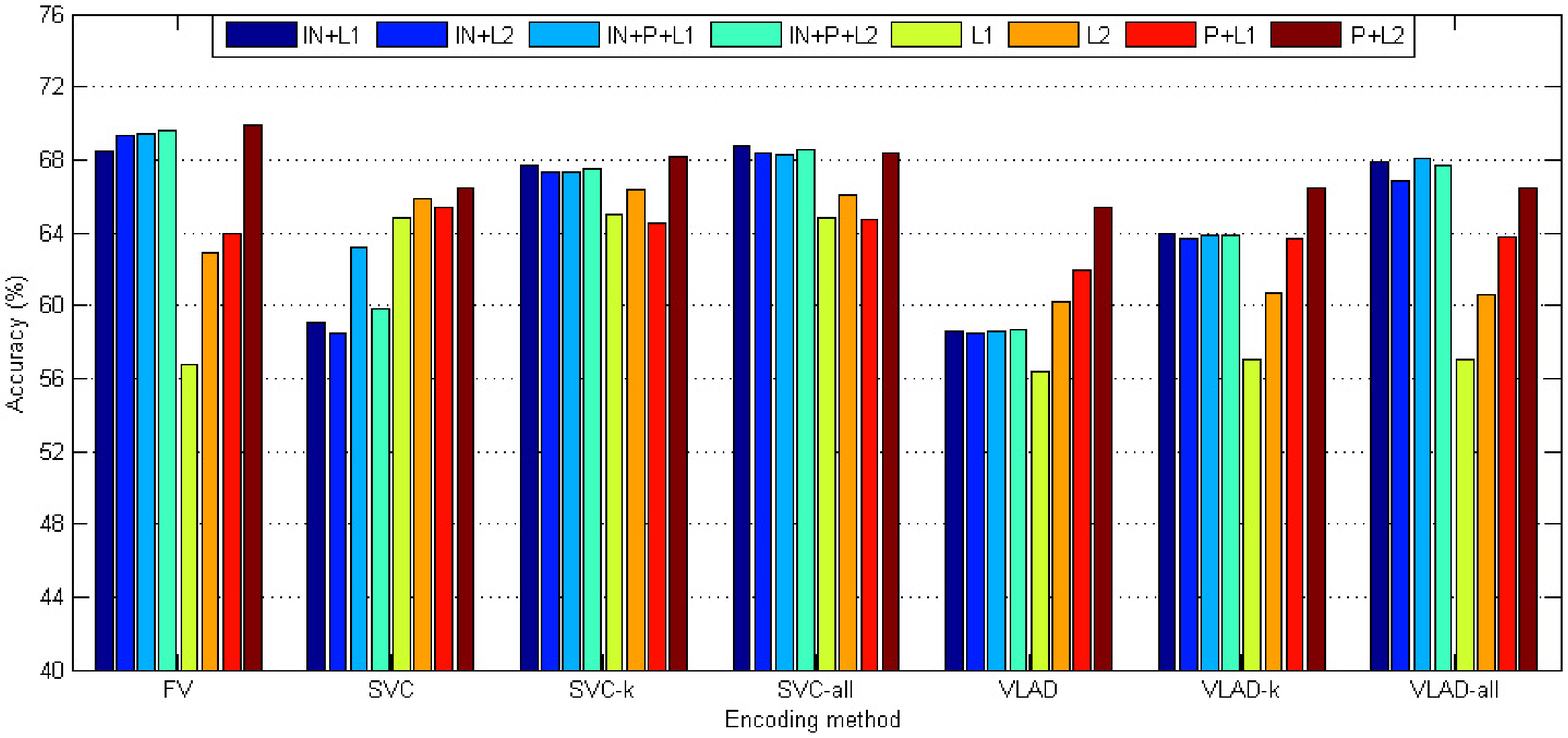}
  \caption{Comparison of different pooling-normalization strategies with \textbf{STIPs} features using descriptor level fusion on the UCF101 dataset. Note that there is only max pooling for voting and reconstruction based encoding methods, and there is only intra normalization for super vector based encoding methods.}
  \label{fig:PoolNorm1}
\end{figure*}
\begin{figure*}
  \center
\includegraphics[width=0.3\linewidth,height=0.3\linewidth]{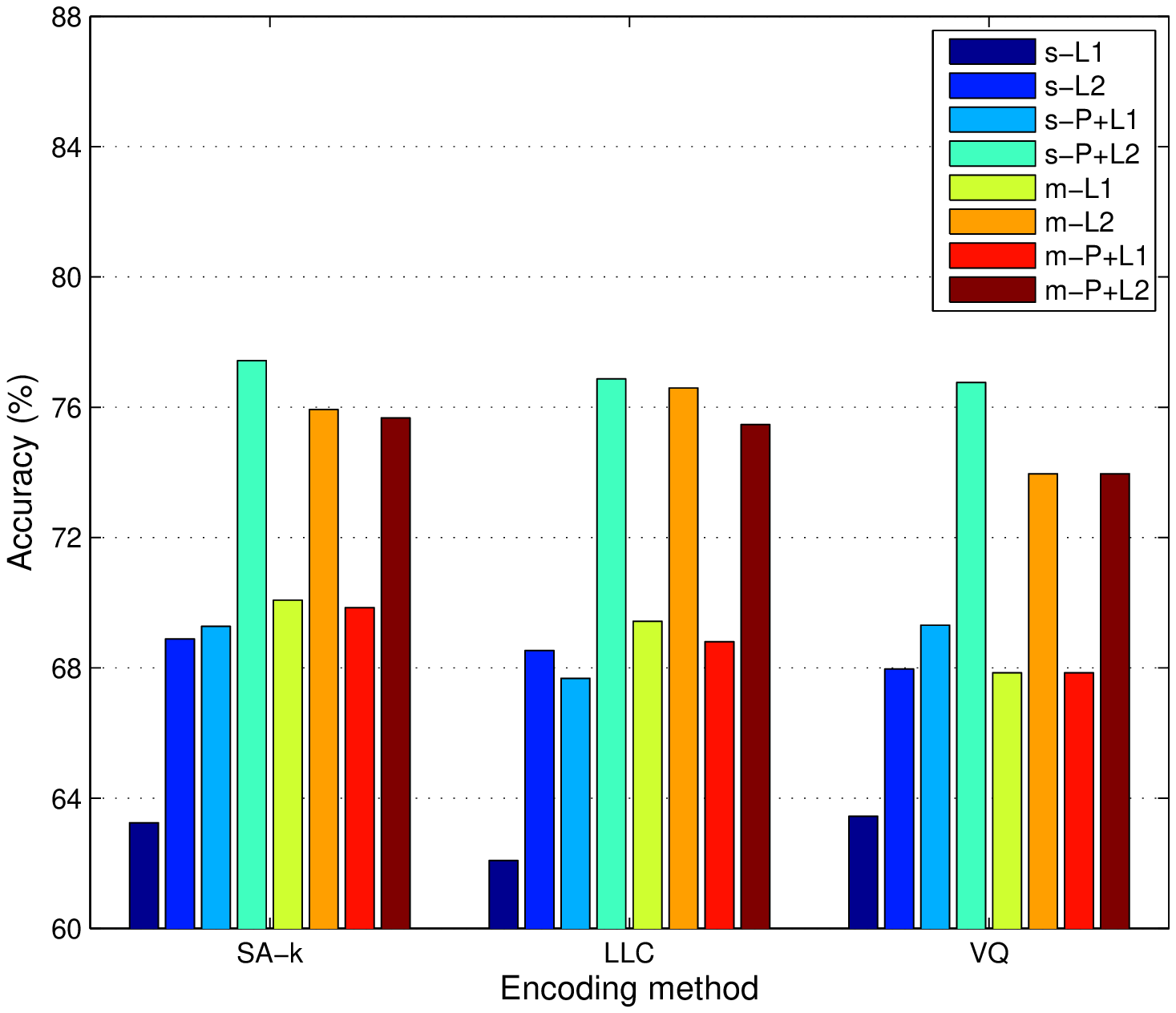}
\includegraphics[width=0.68\linewidth,height=0.3\linewidth]{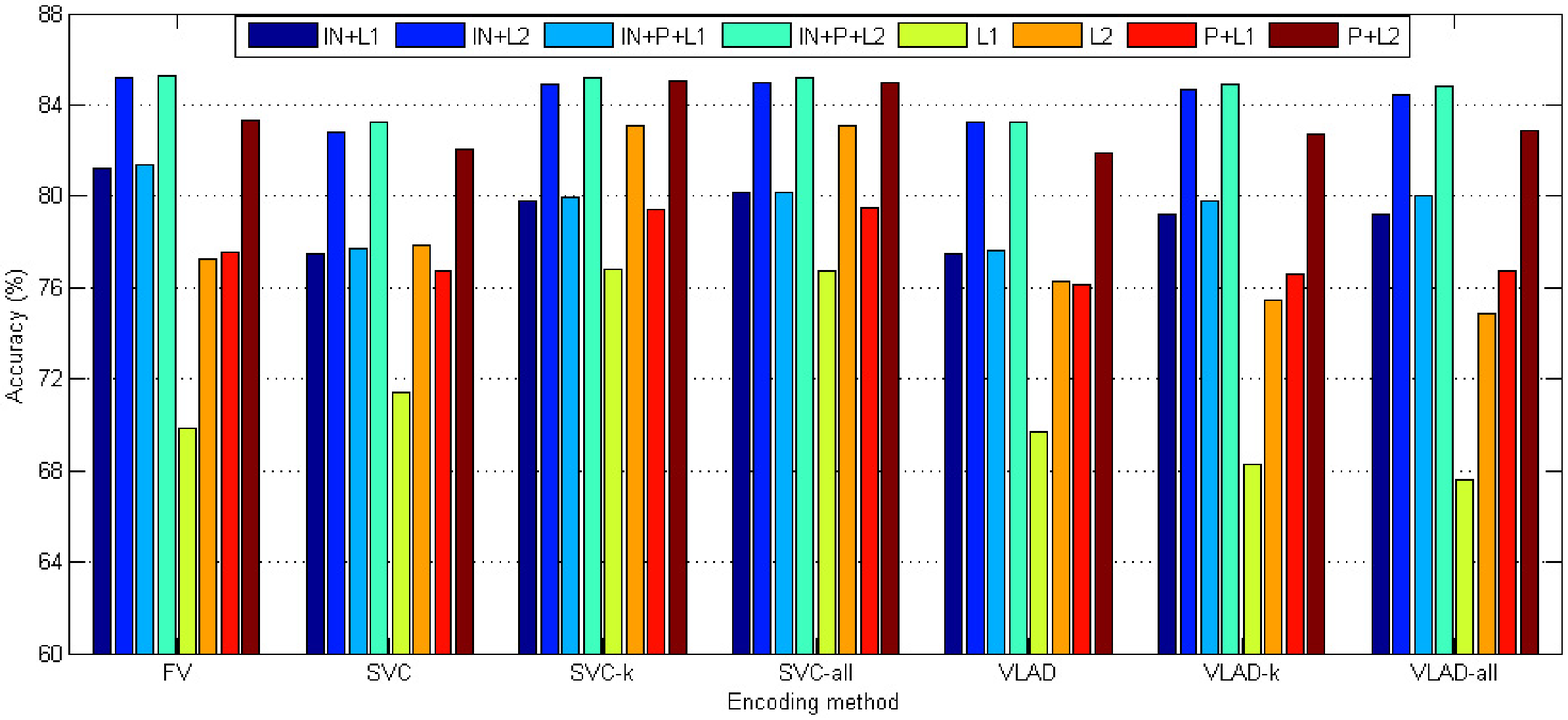}
  \caption{Comparison of different pooling-normalization strategies with \textbf{iDTs} features using descriptor level fusion on the UCF101 dataset. Note that there is only max pooling for voting and reconstruction based encoding methods, and there is only intra normalization for super vector based encoding methods.}
  \label{fig:PoolNorm2}
\end{figure*}

In this section, we mainly investigate the influence on recognition rate for different pooling and normalization strategies on the UCF101 dataset. Based on the performances of different encoding methods on the UCF101 dataset in previous section, we choose the codebook (GMM) size as $512$ for super vector based methods and codebook size as $8,000$ for the other two types of encoding approaches. Meanwhile, according to our conclusion that super vector based encoding is a promising method and soft assignment is an effective way to improve the encoding methods, we extend VLAD to VLAD-$k$ and VLAD-$all$, SVC to SVC-$k$ and SVC-$all$ as described in Section \ref{sec:relation}. Thus, there are totally $10$ kinds of encoding methods.

For super vector based encoding methods, we evaluate eight normalization methods, specified by with or without intra-normalization, with or without power operation, final $\ell_1$ or $\ell_2$ normalization. For the other two types of encoding methods, we choose two pooling methods, namely max pooling and sum pooling, and four normalization methods, namely $\ell_1$-normalization, $\ell_2$-normalization, power $\ell_1$-normalization, and power $\ell_2$-normalization. In our evaluation, the parameter $\alpha$ in power normalization is set as $0.5$.

The experimental results are shown in Figure \ref{fig:PoolNorm1} and Figure \ref{fig:PoolNorm2}. Several observation can be concluded from these results:
\begin{itemize}
  \item For super vector based encoding methods, intra-normalization is an effective way to balance the weight of different codewords and suppress the burst of features corresponding to background. We found this technique works very well when dense features are chosen. A large number of features in iDTs are irrelevant with the action class and intra-normalization can suppress this influence. However, for sparse features, the effect of intra normalization is not so evident, and even cause performance degradation in the case of hard assignment such as VLAD, SVC. We ascribe  this phenomenon to the fact that the STIPs features are usually located in the moving foreground and related with action class. Thus, these descriptors only vote for a subset of codewords, that are highly related with action class. In this case, intra-normalization can decrease the discriminative power of action-related codewords and increase the influence of irrelevant codewords. \textbf{In conclusion, intra-normalization is effective in handling burst of irrelevant features in the case of dense-sampling strategy}.
  \item For different encoding methods and local features, we observe that the final $\ell_2$-normalization outperforms $\ell_1$-normalization. In fact, the normalization method is related to kernel used in final classifier. In our case of linear SVM, the kernel is $k(\mathbf{x},\mathbf{y}) = \mathbf{x}^\top\mathbf{y}$. The choice of $\ell_2$-normalization can ensure two things: (i) $k(\mathbf{x},\mathbf{x}) = \mathrm{const.}$; (ii) $k(\mathbf{x},\mathbf{x}) \geq k(\mathbf{x},\mathbf{y})$. This can guarantee a simple consistency criterion: by interpreting $k(\mathbf{x},\mathbf{y})$ as a similarity score, $\mathbf{x}$ should be the most similar point to itself \cite{VedaldiZ12}. However, the choice of $\ell_1$-normalization can not make sure that the point is most similar to self and may cause the instability during SVM training. \textbf{Above all, $\ell_2$-normalization generally outperforms $\ell_1$-normalization when using linear SVM}.
  \item The influence of power operation in normalization is highly related with pooling method. We observe that power normalization is an effective approach to boost the performance of representation obtained from sum pooling, such as super vector based representation, LLC, SA-$k$ with sum pooling. However, power normalization have little effect for max pooling and sometimes even cause the performance degradation for LLC, SA-$k$. The operation of power usually reduces the difference between different codewords, which means smoothing the histogram. This smooth effect can reduce the influence of high frequent codeword on the kernel calculation and improve the influence of less frequent codeword. For sum pooling, the resulting histogram is usually very sharp and unbalanced due to feature burst, and the smooth operation has a positive effect for suppress the high frequent codeword. However, for max pooling, the histogram is itself not so sharp as sum pooling, and thus the power normalization may have a side effect. \textbf{In above, power operation is an effective strategy to smooth the resulting histogram and can greatly improve the performance of sum pooling representation}.
  \item Among different choices of pooling operations and normalization methods, we conclude that sum pooling and power $\ell_2$-normalization is the best combination. For dense features, intra normalization is an extra bonus for performance boosting. For sparse features, intra normalization sometimes may have a negative effect. The success of sum pooling and power $\ell_2$-normalization can be explained by the Hellinger's kernel \cite{VedaldiZ12}, which has turned out to be an effective kernel to calculate the similarity between two histograms. The linear kernel calculation in the feature space resulting from power $\ell_2$-normalization is equivalent to the Hellinger's kernel calculation in the original space just using $\ell_1$ normalization:
      \begin{equation}
        <\frac{\sqrt{\mathbf{x}}}{\|\sqrt{\mathbf{x}}\|_2},\frac{\sqrt{\mathbf{y}}}{\|\sqrt{\mathbf{y}}\|_2}> = <\sqrt{\frac{\mathbf{x}}{\|\mathbf{x}\|_1}},\sqrt{\frac{\mathbf{y}}{\|\mathbf{y}\|_1}}>
      \end{equation}which means power $\ell_2$-normalization explicitly introduces non-linear kernel in the final classifier. \textbf{In a word, sum pooling and power $\ell_2$-normalization is effective and efficient way to enable linear SVM to have the power of non-linear classifier and boost final recognition rate}.
\end{itemize}

In conclusion, pooling and normalization is a crucial step in the pipeline of BoVW framework, whose importance may not be highlighted in previous research work. Proper choice of pooling and normalization strategy may largely reduce the performance gap of different encoding methods. For sum pooling and power $\ell_2$-normalization, which is the best combination in all these possible choices, the performances of LLC, SA-$k$, and VQ are comparable to each other for iDT features. Thus, in the remaining evaluation for fusion methods, we fix the pooling and normalization strategy as sum pooling and power $\ell_2$-normalization.

\subsection{Exploration of Fusion Methods}
\begin{table*}
  \caption{Comparison of different fusion methods for the encoding methods on the \textbf{HMDB51} dataset.}
  \label{tbl:hmdb}
  \centering
  \begin{tabular}{lcccccccccc}
  \hline
  \hline
  Methods & ~~~FV~~~ & ~~SVC~~ & ~SVC-$k$~ & ~SVC-$all$~ & ~VLAD~ & VLAD-$k$ & VLAD-$all$ & ~~LLC~~ & SA-$k$ & ~~~VQ~~~ \\
  \hline
  \hline
  \multicolumn{11}{l}{\textbf{Space Time Interest Points (STIPs)}} \\
  ~HOG & 22.81 & 17.76 & 21.09 & 21.87 & 18.13 & 19.87 & 20.04 & 20.46 & 18.39 & 16.10 \\
  ~HOF & 31.96 & 30.44 & 32.68 & 33.36 & 30.46 & 31.53 & 31.55 & 27.19 & 26.27 & 24.49\\
  ~d-Fusion & \textbf{38.82} & \textbf{35.12} & 36.64 & \textbf{37.19} & \textbf{34.81} & \textbf{36.18} & \textbf{36.23} & 29.87 & 28.13 & 25.66 \\
  ~r-Fusion & 37.32 & 34.36 & \textbf{36.73} & \textbf{37.19} & 34.23 & 35.84 & 35.88 & \textbf{33.44} & \textbf{32.59} & \textbf{30.35} \\
  ~s-Fusion & 36.71 & 32.14 & 34.51 & 34.99 & 32.11 & 33.90 & 34.01 & 32.52 & 30.96 & 27.54 \\
  \hline
  \multicolumn{11}{l}{\textbf{Improved Dense Trajectories (iDTs)}} \\
  ~HOG & 45.12 & 36.93 & 39.32 & 38.10 & 36.93 & 39.30 & 37.08 & 37.08 & 35.45 & 34.81 \\
  ~HOF & 50.70 & 47.70 & 49.00 & 48.00 & 47.70 & 49.00 & 45.80 & 42.20 & 42.70 & 42.10 \\
  ~MBHx & 44.14 & 39.35 & 43.01 & 41.68 & 39.43 & 43.03 & 41.55 & 35.51 & 35.51 & 34.6 \\
  ~MBHy & 50.04 & 44.25 & 47.02 & 46.51 & 44.27 & 47.02 & 44.68 & 40.39 & 40.35 & 39.78 \\
  ~d-Fusion & 58.37 & 54.12 & 56.82 & 56.86 & 54.2 & 56.88 & 54.73 & 48.25 & 48.58 & 47.93 \\
  ~r-Fusion & \textbf{60.22} & \textbf{58.19} & \textbf{60.09} & \textbf{60.07} & \textbf{58.26} & \textbf{60.09} & \textbf{58.58} & \textbf{55.45} & \textbf{55.8} & \textbf{55.27} \\
  ~s-Fusion &  59.62 & 57.27 & 59.11 & 58.78 & 57.14 & 59.17 &  57.54 & 53.68 & 53.94 & 53.27 \\
  \hline
  \hline
  \end{tabular}
\end{table*}

\begin{table*}
  \caption{Comparison of different fusion methods for the encoding methods on the \textbf{UCF50} dataset.}
  \label{tbl:ucf50}
  \centering
  \begin{tabular}{lcccccccccc}
  \hline
  \hline
  Methods & ~~~FV~~~ & ~~SVC~~ & ~SVC-$k$~ & ~SVC-$all$~ & ~VLAD~ & VLAD-$k$ & VLAD-$all$ & ~~LLC~~ & SA-$k$ & ~~~VQ~~~ \\
  \hline
  \hline
  \multicolumn{11}{l}{\textbf{Space Time Interest Points (STIPs)}} \\
  ~HOG & 66.20 & 60.76 & 63.98 & 63.94 & 60.22 & 62.32 & 62.22 & 60.42 & 59.11 & 56.21 \\
  ~HOF & 73.10 & 71.93 & 74.14 & 74.56 & 71.30 & 72.36 & 72.51 & 64.72 & 63.80 & 61.55\\
  ~d-Fusion & \textbf{78.32} & \textbf{76.33} & 77.60 & 77.59 & \textbf{75.57} & \textbf{76.06} & \textbf{76.13} & 70.13 & 68.66 & 67.16 \\
  ~r-Fusion & 77.21 & 76.07 & \textbf{78.42} & \textbf{78.91} & 75.36 & 75.95 & 75.99 & \textbf{74.05} & \textbf{73.67} & \textbf{71.95} \\
  ~s-Fusion & 76.33 & 76.19 & 77.76 & 77.25 & 73.79 & 74.91 & 74.98 & 72.95 & 71.66 & 69.16 \\
  \hline
  \multicolumn{11}{l}{\textbf{Improved Dense Trajectories (iDTs)}} \\
  ~HOG & 84.39 & 78.22 & 80.29 & 79.97 & 78.19 & 80.20 & 78.33 & 72.73 & 73.76 & 74.27 \\
  ~HOF & 86.33 & 85.18 & 85.92 & 84.94 & 85.15 & 85.87 & 83.48 & 80.23 & 80.58 & 80.29 \\
  ~MBHx & 84.03 & 81.33 & 83.19 & 82.46 & 81.28 & 83.12 & 81.16 & 77.77 & 77.91 & 77.04 \\
  ~MBHy & 87.02 & 84.64 & 86.38 & 85.29 & 84.60 & 86.32 & 84.04 & 80.36 & 80.6 & 80.3 \\
  ~d-Fusion & 90.84 & 89.39 & 90.72 & 90.62 & 89.43 & 90.64 & 90.18 & 84.18 & 84.76 & 84.67 \\
  ~r-Fusion & \textbf{92.07} & \textbf{90.87} & \textbf{91.89} & \textbf{91.50} & \textbf{90.82} & \textbf{91.80} & \textbf{90.56} & \textbf{87.56} & \textbf{87.92}  & \textbf{88.12} \\
  ~s-Fusion &  91.03 &  90.08 & 90.71 & 90.36 & 90.11 & 90.63 &  89.67 &   87.37 & 87.86 & 87.41 \\
  \hline
  \hline
  \end{tabular}
\end{table*}

\begin{table*}
  \caption{Comparison of different fusion methods for the encoding methods on the \textbf{UCF101} dataset.}
  \label{tbl:ucf101}
  \centering
  \begin{tabular}{lcccccccccc}
  \hline
  \hline
  Methods & ~~~FV~~~ & ~~SVC~~ & ~SVC-$k$~ & ~SVC-$all$~ & ~VLAD~ & VLAD-$k$ & VLAD-$all$ & ~~LLC~~ & SA-$k$ & ~~~VQ~~~ \\
  \hline
  \hline
  \multicolumn{11}{l}{\textbf{Space Time Interest Points (STIPs)}} \\
  ~HOG & 53.74 & 47.56 &  50.07 &  50.31 & 47.15 & 49.21 & 49.35 & 46.70 & 45.79 & 42.85 \\
  ~HOF & 62.89 & 60.57 &  63.81 &  64.02 & 60.04 & 61.73 & 61.60 & 54.16 & 52.78 & 50.04\\
  ~d-Fusion & \textbf{69.90} & \textbf{66.43} & 68.22 &  68.40 & \textbf{65.42} & \textbf{66.42} & \textbf{66.46} & 59.52 & 57.83 & 56.09 \\
  ~r-Fusion & 68.21 & 65.39 & \textbf{69.00} & \textbf{69.18} & 65.39& 66.13 & 66.19 & \textbf{63.04} & \textbf{62.13} & \textbf{59.31} \\
  ~s-Fusion & 66.77 & 62.50 & 65.81 & 65.98 & 62.17 & 63.97 & 64.15 & 60.94 & 59.48 & 56.69 \\
  \hline
  \multicolumn{11}{l}{\textbf{Improved Dense Trajectories (iDTs)}} \\
  ~HOG & 74.79 & 69.74 & 72.14 & 72.36 & 69.66 & 71.65 & 71.39 & 65.46 & 65.81 & 65.40 \\
  ~HOF & 78.63 & 76.26 & 77.70 & 77.12 & 76.28 & 77.76 & 76.35 & 71.03 & 71.14 & 70.57 \\
  ~MBHx & 76.82 & 71.63 & 74.24 & 73.92 & 71.62 & 74.11 & 71.84 & 67.00 & 67.55 & 66.43 \\
  ~MBHy & 79.15 & 74.53 & 77.46 & 76.82 & 74.54 & 76.78 & 74.21 & 69.6 & 69.67 & 68.50 \\
  ~d-Fusion & 85.32 & 83.36 & 85.19 & 85.17 & 83.39 & 85.14 & 85.45 & 77.65 & 77.96 & 76.76 \\
  ~r-Fusion & \textbf{87.11} & \textbf{84.87} & \textbf{86.54} & \textbf{86.19 }& \textbf{84.90} & \textbf{86.16} & \textbf{85.59} & \textbf{81.43} & \textbf{81.65} & \textbf{81.37} \\
  ~s-Fusion &  85.49 & 83.34 & 84.84 & 84.57 & 83.29 &  85.04 & 83.83 &  80.11 &  80.39 & 79.81 \\
  \hline
  \hline
  \end{tabular}
\end{table*}

The local features usually have multiple descriptors, such as HOG, HOF, MBHx, and MBHy, each of which corresponds to a specific view of video data. For the empirical study in previous section, we choose a simple method to combine these multiple descriptors, where we just concatenate them into a single one, namely descriptor level fusion. In this section, we mainly analyze the influence of different fusion methods on final recognition performance.

For encoding methods, we choose the same ten approaches as in previous section. The codebook size of super vector based methods is set as $512$ and the one of other encoding methods is set as $8,000$. For pooling and normalization methods, we use sum pooling and power $\ell_2$-normalization, according to the observations in Section \ref{sec:poolingnorm}. We also use intra normalization for super vector based encoding methods of iDTs features. For fusion methods, we evaluate three kinds of methods, namely descriptor level fusion, representation level fusion, and score level fusion, as described in Section \ref{sec:fusion}. For score level fusion, we use the geometrical mean to combine the scores from multiple SVMs.

\begin{table*}
  \caption{Comparison our hybrid representation with the sate-of-the-art methods.}
  \label{tbl:cmp}
  \centering
  \begin{tabular}{lcc|lcc|lcc}
  \hline
  \hline
  HMDB51 & ~Year~ & \% & ~UCF50 & ~Year~ & \% & ~UCF101 & ~Year~ &\% \\
  \hline
  \hline
  Kuehne \emph{et al.} \cite {KuehneJGPS11} & 2011 & 23.0 &  Sadanand \emph{et al.} \cite{SadanandC12} & 2012 & 57.9 & Soomro \emph{et al.} \cite{SOOMRO12} & ~2012~ & 43.9 \\
  Sadanand \emph{et al.} \cite{SadanandC12} & 2012 & 26.9 & Kliper-Gross \emph{et al.} \cite{Kliper-GrossGHW12} & 2012 & 72.7 & Karpthy \emph{et al.} \cite{Karpathy14} & 2014 & 63.3 \\
  Kliper-Gross \emph{et al.} \cite{Kliper-GrossGHW12} & 2012 & 29.2 & Solmaz \emph{et al.} \cite{SolmazAS13} &  2012 & 73.7  &  Cai \emph{et al.} \cite{CaiWPQ14} & 2014 & 83.5  \\
  Jiang \emph{et al.} \cite{JiangDXLN12} & 2012 & 40.7 & Reddy \emph{et al.} \cite{ReddyS13} & 2012 & 76.9  & Wu \emph{et al.} \cite{WuZL14} & 2014 & 84.2 \\
  Wang \emph{et al.} \cite{WangQT13} & 2013 & 42.1  & Wang \emph{et al.} \cite{WangQT13} & 2013 & 78.4 &  Peng \emph{et al.} \cite{PengWCQP13} & 2013 & 84.2  \\
  Wang \emph{et al.} \cite{WangKSL13} & 2013 & 46.6 &  Wang \emph{et al.} \cite{WangKSL13}  & 2013 & 84.5 & Murthy \emph{et al.} \cite{Murthy13} & 2013 & 85.4   \\
  Peng \emph{et al.} \cite{PengQPQ13}& 2013 & 49.2 & Wang \emph{et al.} \cite{WangQT13b}  & 2013 & 85.7  & Karaman \emph{et al.} \cite{Karaman13} & 2013 & 85.7 \\
  Wang \emph{et al.} \cite{WangS13a} & 2013 & 57.2 & Wang \emph{et al.} \cite{WangS13a} & 2013 & 91.1 & Wang \emph{et al.} \cite{Wang13} & 2013 & 85.9 \\
  \hline
  Hybrid representation & - & \textbf{61.1} & Hybrid representation & - & \textbf{92.3} & Hybrid representation & - & \textbf{87.9} \\
  \hline
  \hline
  \end{tabular}
\end{table*}

The experimental results on three datasets are shown in Table \ref{tbl:hmdb}, Table \ref{tbl:ucf50}, and Table \ref{tbl:ucf101}. From these results, we observe serval trends:
\begin{itemize}
  \item \textbf{For iDTs features, representation level fusion is the best choice for all of the selected encoding methods on the three datasets}. This result indicates that these multiple descriptors are most correlated in the video level. Descriptor level fusion emphasizes the dependance in cuboid and results in high dimension features for codebook training and encoding. This may make these unsupervised learning algorithm unstable.
  \item \textbf{For STIPs features, representation level fusion is more effective for reconstruction based and voting based encoding methods}. For super vector based encoding methods, the performance of representative level fusion is comparable to that of descriptor level fusion. This trend is consistent with the finds with iDTs features.
  \item \textbf{For both features, SA-$k$, LLC, and VQ encoding methods are much sensitive to fusion methods than those super vector based encoding methods}. Great improvement can be obtained for SA-$k$, LLC, and VQ when using representation level fusion, but slight improvements happen to those super vector methods. We analyze this is due to two facts. Firstly, for reconstruction and voting based encoding methods, the final dimension of representation level fusion is $M$ (the number of descriptors) times of the dimension of descriptor level fusion. However, for super vector based encoding methods, the dimension of descriptor level fusion is the same with representation level fusion. The higher dimension of final representation may enable SVM to classify more easily. Secondly, the codebook size $K$ of super vector methods is much smaller than that of other types of encoding methods, where clustering algorithm may be more stable for high dimensionality in descriptor level fusion method.
\end{itemize}

Based on the observation and analysis above, we conclude that fusion method is a very important component for handling combination of multiple descriptors in the action recognition system. Representation level fusion method is a suitable choice for different kinds of encoding methods due to its good performance. From our analysis, we know that the performance boosting of fusing multiple features mainly owns the complementarity of these features. This complementarity may be not limited to the exploration of different descriptors, but also can be extended to the different BoVWs. From the perspective of statistics, FV aggregates information using $1^{st}$ and $2^{nd}$ order statistics, while SVC is about zero and $1^{st}$ order statistics. Intuitively, these two kinds of super vector encoding methods are complementary to each other. Thus, we present a new feature representation, called hybrid representation, combining the outputs FV and soft version SVC of multiple descriptors, including HOG, HOF, MBHx, and MBHy. This representation is simple but proved to be effective in next section.

\subsection{Comparison to the State-of-the-Art Results}
\label{sec:stoa}
In this section, we demonstrate the effectiveness of our proposed hybrid representation according to our previous insightful analysis. Specifically, we choose two super vector based encoding methods, namely SVC-$k$ and FV, for iDTs features. We use the power operation and then intra $\ell_2$-normalization. For feature fusion, we adopt the representation level fusion method.

Table \ref{tbl:cmp} shows our final recognition rates and compare our results to that of state-of-the-art approaches. For the HMDB51 dataset, we obtain a recognition rate of $61.1\%$, which is superior to the best result \cite{WangS13a} by $3.9\%$. Our system reaches classification accuracy of $92.3\%$ on the dataset of UCF50 and $87.9\%$ on the dataset of UCF101, which outperform the best results by $1.2\%$ and $2.0\%$ respectively. It is worth noting that UCF101 is newest and largest dataset, so few published papers have reported results on this dataset. We mainly compare with those top performers in the Thumos'13 Action Recognition Challenge \cite{THUMOS13}. We also compare with three latest papers in CVPR 2014. Karpathy \emph{et al.} \cite{Karpathy14} resorts to a large deep Convolutional Neural Network trained with an extra 1-M training dataset. Cai \emph{et al.} \cite{CaiWPQ14} propose a complex and less efficient encoding method by considering the correlation of different descriptors. Wu \emph{et al.} \cite{WuZL14} propose a simple, lightweight, but powerful bimodal encoding method. Our results outperform these top performer and latest papers on the UCF101. From these comparisons, our hybrid representation is an efficient and effective method and obtains the state-of-the-art performance on the three challenging datasets.

\section{Conclusion}
\label{sec:conclusion}
In this paper, we have comprehensively studied each step in the BoVW pipeline and tried to uncover good practice to build a more accurate and efficient action recognition system. Specifically, we mainly explore five aspects, namely local features, pre-processing techniques, encoding methods, pooling and normalization strategy, fusion methods. We conclude that every step is crucial for contributing to the final recognition rate and improper choice in one of the steps may counteract the performance improvement of other steps. Meanwhile, based on the insights from our comprehensive study, we propose a simple yet effective representation, called \emph{hybrid representation}. Using this representation, our action recognition system obtains the state-of-the-art performance on the three challenging datasets.

%\begin{acknowledgements}
%\end{acknowledgements}

% \bibliographystyle{splncs}
% \bibliographystyle{spmpsci}
\bibliographystyle{spmpsci}      % mathematics and physical sciences
\bibliography{survey}   % name your BibTeX data base

\end{document}